\documentclass[10pt,twocolumn,letterpaper]{article}

\usepackage[accsupp]{axessibility}
\usepackage[pagenumbers]{cvpr}

\usepackage{graphicx}
\usepackage{amsmath}
\usepackage{amssymb}
\usepackage{amsthm}
\usepackage[ruled,linesnumbered]{algorithm2e}

\newtheorem{prop}{Proposition}
\usepackage{booktabs}
\usepackage{caption}
\usepackage{subcaption}
\usepackage{xcolor}
\usepackage{multirow}

\usepackage{xspace}
\usepackage{kotex}

\newcommand{\methodname}{CGL\xspace}
\newcommand{\methodnamefull}{\textbf{C}onfidence-based \textbf{G}roup \textbf{L}abel assignment (\textbf{\methodname})\xspace}
\newcommand{\ours}{\methodname}

\newcommand{\ourscenariofull}{Algorithmic Group \textbf{Fair}ness with the \textbf{P}artially annotated \textbf{G}roup labels (\textbf{\ourscenario})\xspace}
\newcommand{\ourscenario}{Fair-PG\xspace}

\newcommand{\widebar}[1]{\mkern 1.5mu\overline{\mkern-1.5mu#1\mkern-1.5mu}\mkern 1.5mu}

\usepackage[pagebackref,breaklinks,colorlinks]{hyperref}

\usepackage[capitalize]{cleveref}
\crefname{section}{Sec.}{Secs.}
\Crefname{section}{Section}{Sections}
\Crefname{table}{Table}{Tables}
\crefname{table}{Tab.}{Tabs.}

\begin{document}

\title{Learning Fair Classifiers with Partially Annotated Group Labels}

\author{Sangwon Jung$^1$\thanks{Works done while doing an internship at NAVER AI Lab.} ~~ Sanghyuk Chun$^2$\thanks{Corresponding authors} ~~ Taesup Moon$^{1,3}$\footnotemark[2]\vspace{.1in}\\
$^1$ Department of ECE/ASRI, Seoul National University \quad $^2$ NAVER AI Lab \\ $^3$ Interdisciplinary Program in Artificial Intelligence, Seoul National University}
\maketitle

\begin{abstract}
Recently, fairness-aware learning have become increasingly crucial, but most of those methods operate by assuming the availability of fully annotated demographic group labels. We emphasize that such assumption is unrealistic for real-world applications since group label annotations are expensive and can conflict with privacy issues. In this paper, we consider a more practical scenario, dubbed as Algorithmic Group \textbf{Fair}ness with the \textbf{P}artially annotated \textbf{G}roup labels (\textbf{Fair-PG}). We observe that the existing methods to achieve group fairness perform even worse than the vanilla training, which simply uses full data only with target labels, under Fair-PG. To address this problem, we propose a simple \textbf{C}onfidence-based \textbf{G}roup \textbf{L}abel assignment (\textbf{CGL}) strategy that is readily applicable to any fairness-aware learning method. CGL utilizes an auxiliary group classifier to assign pseudo group labels, where random labels are assigned to low confident samples. We first theoretically show that our method design is better than the vanilla pseudo-labeling strategy in terms of fairness criteria. Then, we empirically show on several benchmark datasets that by combining CGL and the state-of-the-art fairness-aware in-processing methods, the target accuracies and the fairness metrics can be jointly improved compared to the baselines. Furthermore, we convincingly show that CGL enables to naturally augment the given group-labeled dataset with external target label-only datasets so that both accuracy and fairness can be improved. Code is available at \url{https://github.com/naver-ai/cgl_fairness}.
\end{abstract}

\section{Introduction}
\label{sec:intro}

\begin{figure}[t]
    \centering
    \includegraphics[width=\linewidth]{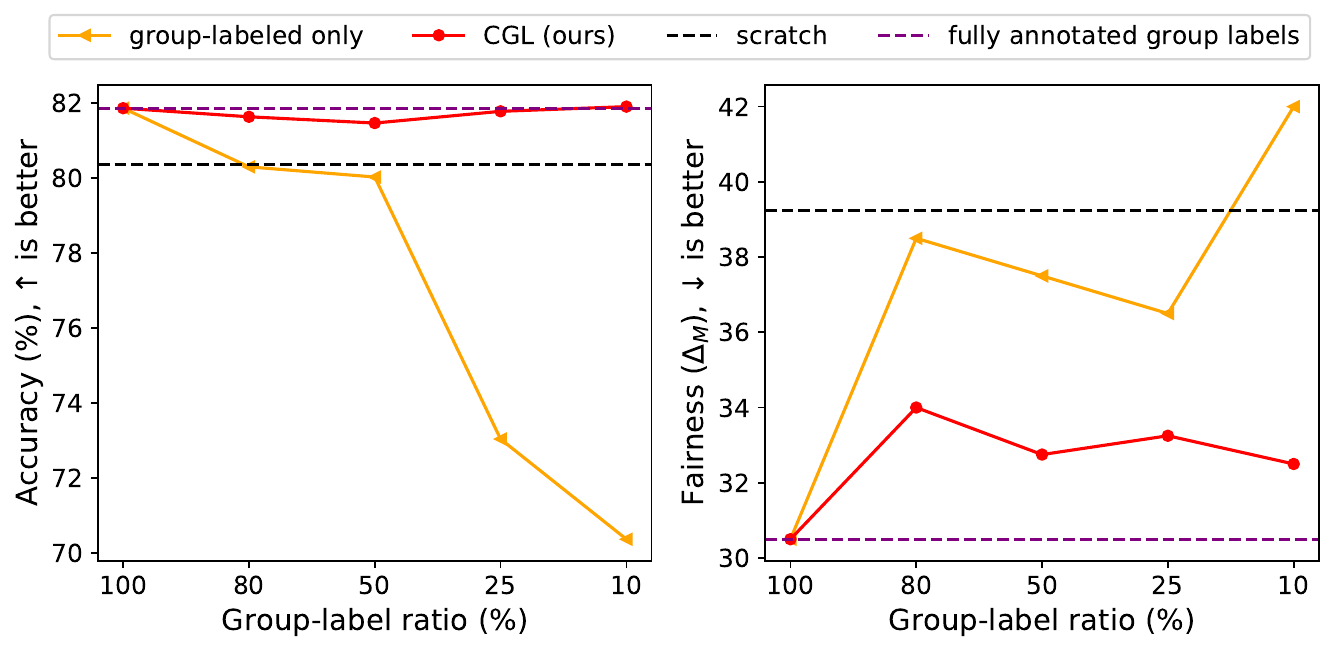}
    \caption{\small \textbf{Can fair-training methods still learn fair classifiers when group labels are partially annotated?}
    We note the state-of-the-art fairness fair-training FairHSIC \cite{quadrianto2019discovering} using only the group-labeled subset (yellow) shows worse fairness criterion ($\Delta_M$, \cref{eq:deltam}, lower the better) than the ``scratch'' (\ie, no consideration of a fairness criteria) in the low group label regime (\eg, 10\%) on UTKFace \cite{utkface}.
    Our \ours (red), on the other hand, can be potentially applied to any fair-training method, and when it is combined with FairHSIC, both the target accuracy and the fairness criteria are significantly improved for the low group label regime. 
    }
    \label{fig:teaser}
    \vspace{-1em}
\end{figure}

Recent advances of machine learning (ML) models have witnessed promising outcomes even in societal applications, such as credit estimation \cite{khandani2010consumer}, crime assessment systems \cite{brennan2009evaluating, COPMAS}, automatic job interviews \cite{nguyen2016hirability}, face recognition \cite{buolamwini2018gender, wang2019racial}, and law enforcement \cite{garvie2016perpetual}. However, machines are often more inaccurate to a particular group (\eg, darker-skinned females) than other groups (\eg, lighter-skinned males) \cite{buolamwini2018gender}, \ie, machines are discriminatory.
To mitigate the issue, \textit{fairness-aware learning} has recently emerged; a model should not discriminate against any demographic group with sensitive attributes, \eg, age, gender, or race.

Many existing approaches for \textit{group fairness} \cite{zemel2013learning, zafar2017fairness, agarwal2018reductions, quadrianto2019discovering, jiang2020identifying, chuang2021fair, jung2021mfd} utilize two types of labels: \textit{target labels}, which are task-oriented (\eg, crime assessment) and \textit{group labels}, which are defined by socially sensitive attribute groups (\eg, ethnicity or gender).
Many existing methods for achieving group fairness rely on the group labels to train fair classifiers. For example, many approaches explicitly minimize the statistical parity metrics between groups defined by sensitive attributes. However, in many realistic applications, \textit{e.g.,} computer vision, assuming that all images have sensitive group labels can be unrealistic and make the existing methods impractical. First, in many image datasets, group labels are not explicitly given as in tabular datasets \cite{brennan2009evaluating, khandani2010consumer, COPMAS} but are defined in high-level semantics, requiring additional expensive human annotations. Secondly, the sensitive attributes are usually personal information protected by laws, such as EU General Data Protection Regulation (GDPR). Hence, in real-world applications, collecting group labels for all data points are impossible without permissions by all users and, furthermore, sensitive attributes cannot be persistently stored but should be expired. Thus, the underlying assumption by the previous fair-training methods, \ie, group labels are fully annotated, can limit their usability in real-world applications.\vspace{-.1in}

\paragraph{Contribution.}
In this work, we propose and investigate a less explored but very practical problem: \textit{\ourscenariofull}.
Many existing fair-training methods for group fairness assume all training samples have group labels, and optimize fairness constraints by the group labeled training samples. In this case, they cannot be directly applied to the \ourscenario problem.
We empirically show that the baseline fair-training methods, which operate only on the group-labeled samples, perform even worse than the vanilla ``scratch'' training that use all the training samples,
in terms of fairness when the number of group-labeled samples is small (\eg, 10\%) -- See \cref{fig:teaser}.
Although there exist a few attempts to achieve algorithmic fairness without demographics labels \cite{hashimoto2018fairness, lahoti2020fairness}, they do not directly solve the \textit{group fairness} problem. Also, they do not utilize partially annotated group labels at all, while a small number of labeled data can improve the overall performances.
To this end, we propose a simple yet effective strategy for \ourscenario that can be applied to \textit{any} fair-training methods for group fairness, dubbed as \methodnamefull.
\ours assigns pseudo group labels to group-unlabeled samples using an auxiliary group classifier, if the predictions are sufficiently confident, and random group labels, otherwise.

We provide high-level understandings of how \methodname works on the \ourscenario scenario. We theoretically support that (1) the fairness parity computed by our approach approximates the parity of the underlying group label distribution better than the one by the vanilla pseudo-label strategy which totally trusts the predictions of the auxiliary group classifier, (2) assigning a random group label to a data point implies the elimination of the fairness constraint of the sample. In practice, since the existing fair-training methods use a relaxed constraint, \methodname can be interpreted as a regularization method for the low confident group-unlabeled samples.

In our experiments, the combination of \methodname with state-of-the-art fair-traning methods (\eg, MFD \cite{jung2021mfd}, FairHSIC \cite{quadrianto2019discovering} and LBC \cite{jiang2020identifying}) has consistently and significantly improved target accuracies as well as fairness parities even under the low group label regime on facial image  \cite{utkface, celeba} and tabular \cite{COPMAS, dua2017uci} datasets. For example, compared to the ``group-labeled only'' baseline, the combination of \methodname and MFD shows +8.23\% target accuracy increase and -8.75 disparity of equal opportunity (DEO) decrease on UTKFace \cite{utkface}, when only 10\% of data points have group labels.
Further extending this result, by augmenting the full UTKFace training set with extra group-unlabeled dataset in \cite{karkkainen2021fairface}, we show that \ours can significantly improve the performance of MFD by +0.92\% accuracy and -5.5 DEO. This is promising since it shows \ours can improve both the accuracy and fairness of a baseline method by augmenting the training data with target label-only dataset, which is relatively easier to obtain than jointly requiring the group labels.

\section{Related Works}
\label{sec:relwork}

\paragraph{Fair-training for group fairness.} 
There have been various works to tackle fairness problem in machine learning models. At a high level, it can usually be classified into three categories, including 1) \emph{individual fairness} \cite{dwork2012fairness, yurochkin2019training} that aims to treat similar users similarly, 2) \emph{group fairness} \cite{hardt2016equality} of which goal is to reduce the statistical parity between groups defined by the sensitive attributes, 3) \emph{Rawlsian min-max fairness} \cite{lahoti2020fairness, dwork2017decoupled, hashimoto2018fairness} which designs to improve the worst performance among groups. In this paper, we follow the notion of \emph{group fairness} in arguing the fairness of a model.
Many fair-training methods have been developed to achieve group fairness. The fairness methods (for group fairness) can be divided into three categories depending on where the technique for fairness is injected into; \textit{pre-processing} methods \cite{zemel2013learning, quadrianto2019discovering, creager2019flexibly} modify a training dataset before learning a model; \textit{in-processing} \cite{kamishima2012fairness, zafar2017fairness, jung2021mfd, chuang2021fair, adv_debiasing, jiang2020identifying} methods consider fairness during training time; \textit{post-processing} methods \cite{alghamdi2020model} modify a trained model. However, despite technical advances for achieving group fairness, existing methods for group fairness have not considered the setting in which a part of a training dataset lacks group label (\ie, \ourscenario).

\paragraph{Fairness with imperfect sensitive attributes.}
Recently, a few attempts have been proposed to consider algorithmic fairness with imperfect sensitive attributes, \eg, noisy group labels. Chen \etal \cite{chen2019fairness} and Kallus \etal \cite{kallus2021assessing} proposed methods for assessing the disparity when only proxy variables (\eg, surname) for the protected variables are given. Meanwhile, several works \cite{wang2020robust,lamy2019noise} posed solutions of learning a fair classifier robust to \textit{noisy} group labels. However, these approaches focus only on \textit{noisy} group labels, hence they cannot be directly applied to our Fair-PG setting.

There also have been a few works for fair-training without any information of protected attributes. Hashimoto \etal \cite{hashimoto2018fairness} proposed a distributionally robust optimization (DRO) \cite{namkoong2017variance} based approach, and Lahoti \etal \cite{lahoti2020fairness} utilized an adversary to identify regions with high loss and re-weight them. Similarly, there exist \textit{de-biasing} methods to solve the bias problem without any labels denoting bias (discussions given in the Appendix). However, these methods have two limitations compared to \ours on \ourscenario. First, they do not directly aim to achieve \emph{group fairness}, but they consider the \emph{Rawlsian Min-Max fairness} or \emph{cross-bias generalization}. Second, it is not straightforward to combine them to the existing fair training methods, while our method can be universally applicable to any fair-training method.

\paragraph{Semi-supervised learning.}
Semi-supervised learning (SSL) \cite{chapelle2009semi} aims to learn a model with a small number of labeled samples and a large number of unlabeled samples.
Our \ourscenario scenario also considers when group labels are partially annotated but target labels are fully annotated.
However, since the aim for SSL is to simply \emph{predict} the future attribute labels as accurately as possible from the partial annotations in the training set, it is not clear whether the predicted attribute labels can be directly plugged-in to achieve the \emph{group fairness} in the test set.
In addition, the recent state-of-the-art SSL methods \cite{berthelot2019mixmatch, berthelot2019remixmatch, sohn2020fixmatch} are hardly applicable to the fairness problem directly because they mostly focus on seeking better augmentation methods and consistent constraints for the augmented inputs.
Instead, our method is motivated from the pseudo-labeling (PL) strategy \cite{lee2013pseudo} to avoid complex modifications on the base fair-training methods by assigning pseudo group labels to the group-unlabeled samples.
While the original PL fully trusts the network predictions, we set the random labels for the low confident samples. We show, both theoretically and empirically, that such \textit{random label selection strategy} in \ours is critical for achieving better group fairness, and we believe such finding is not straightforward. 
Our strategy is also similar to the recently proposed UPS \cite{rizve2021ups} in the SSL context, which withdraws pseudo-labels for low confident samples by using an external uncertainty prediction module. However, \ours uses the full group-unlabeled samples by the random label strategy and outperforms the UPS-base strategy in our experiments.

\begin{figure*}[t!]
    \centering
    \includegraphics[width=0.73\textwidth]{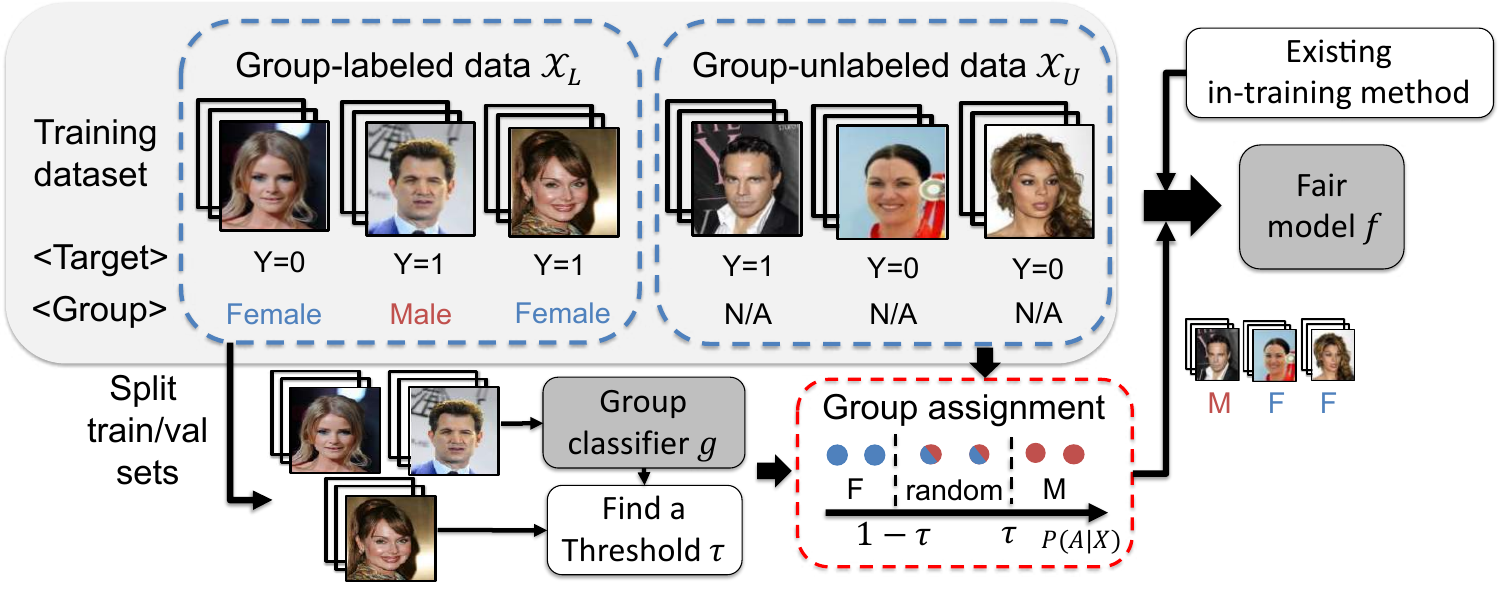}
    \caption{\small {\bf Overview of the proposed \ours strategy under the \ourscenario scenario.} We train a fair model by assigning group pseudo-labels to group-unlabeled training set $\mathcal X_U$. We train an auxiliary group classifier $g$ to generate pseudo labels. Here, we assign random group labels to low confident samples as shown in the dotted red box of the figure ($A=0$ and $A=1$ indicates ``Female'' and ``Male'', respectively). After assigning confidence-based pseudo-labels to the group-unlabeled training set, we apply a fair-training method to train a fair model $f$.}
    \label{fig:main_figure}
    \vspace{-1em}
\end{figure*}

\section{Problem Definition}

In this section, we formally define our scenario, \ourscenario, and the fairness criterion, \textit{disparity of equality of opportunity (DEO)}, for the general $M$-ary classification problem.

\subsection{Formal definition of \ourscenario}
Let $X \in \mathcal{X} \subset \mathbb{R}^d$ be an input feature, $Y \in \mathcal{Y}=\{1, \dots, M\}$ be a target label. We also denote $A \in \mathcal{A}=\{1, \dots, N\}$ as a group label defined by one or multiple sensitive attributes. For example, if phenotype and gender are sensitive attributes, our group labels are $\{$\textit{lighter-skinned male, lighter-skinned female, darker-skinned male} and \textit{darker-skinned female}$\}$. \ourscenario assumes that the input space $\mathcal{X}$ is partitioned into the group-labeled and group-unlabeled sets, $\mathcal{X}_L$ and $\mathcal{X}_U$. That is, a sample $(x,a,y) \sim P(X,A,Y)$ has a group label if $x \in \mathcal{X}_L$, and vice versa if $x \in \mathcal{X}_U$ as illustrated in \cref{fig:main_figure}. With partially annotated group labels, our goal is to find a classifier $f:\mathcal{X}\rightarrow \mathcal{Y}$ not biased against the group label $A$ while predicting a target label that best corresponds to an input feature. 

\subsection{Fairness criterion}

Various group fairness criteria have been proposed with different philosophies of how to define discrimination \cite{dwork2012fairness, hardt2016equality, chouldechova2017fair}.
We consider the \textit{equal opportunity} (EOpp) \cite{hardt2016equality}
for $M$-ary classification problem with non-binary group labels, while most of group fairness criteria assumes binary target or group labels.
A classifier $f$ satisfies EOpp with respect to the sensitive group label $A$ and the target $Y$ if the model prediction $\tilde{Y}$ and $A$ are conditionally independent given $Y$, \ie, $\forall a, a' \in \mathcal{A}$, $y \in \mathcal{Y}$,  $P(\tilde{Y}=y|A=a,Y=y)=P(\tilde{Y}=y|A=a',Y=y)$.
For measuring the degree of unfairness of $f$ under the distribution $P(X|A,Y)$, we use two types of \textit{disparity of EOpp (DEO)}
upon taking the maximum or the average over $y$ as follows, respectively:
{\small
\vspace{-1em}
\begin{align}
\Delta(f, P, y) := &\Big(\max_{a,a'} \Big(|\mathbb{E}_{P(X|A=a,Y=y)} [\mathbb{I}(f(X)=y)]  \nonumber \\& -\mathbb{E}_{P(X|A=a',Y=y)}[\mathbb{I}(f(X)=y)]|\Big)\Big), \label{eq:delta_base}
\end{align}
\vspace{-1.5em}
\begin{align}
\Delta_M(f,P) \triangleq \max_{y} \Delta(f, P, y), \quad \Delta_A(f,P) \triangleq ~
\frac{1}{|\mathcal{Y}|} \sum_{y \in \mathcal{Y}} \Delta(f, P, y)
\label{eq:deltam}
\end{align}
}%
The above two metrics indicate the accuracy gap between groups given a target label and complement each other in showing the worst case and average accuracy gaps.

\section{Confidence-based Group Label Assignment}
\label{sec:method}

In this section, we present our \methodnamefull, which is simple and readily applicable to any fair-training method for the \ourscenario scenario.
Our method is described in details, and its theoretical understanding is provided as well.

\subsection{Method overview}

As the existing fair-training methods for group fairness explicitly utilize group labels to optimize the fairness constraints, they are not directly applicable to our \ourscenario problem.
A naive approach to apply the existing methods to \ourscenario is to use only group-labeled samples for the training. Unfortunately, as we observed in \cref{fig:teaser} and our experiments, this naive baseline performs even worse than the scratch training method that only uses the target labels, in terms of fairness.
As another baseline, we can employ a pseudo-labeling strategy \cite{lee2013pseudo} that assigns the estimated group labels to the group-unlabeled data by training a separate group classifier.
The vanilla pseudo-labeling strategy enables the recent improvements in algorithmic fairness to be readily transferred into our \ourscenario scenario, other than the scratch training only with the labeled training set.
However, since this vanilla pseudo-labeling strategy trains the group classifier only with the group-labeled training samples, the pseudo group labels can be noisy and incorrect.
Compared to the SSL problem, the incorrect group pseudo labels may lead to a more severe issue in terms of fairness by propagating group label errors into the complex fair-training methods.

To that end, we employ \methodnamefull to reduce the effects of incorrect pseudo-labels. As classifier confidences can be a proxy measure of the mis-classification for the given samples \cite{guo2017calibration, hendrycks2016baseline}, we assume that the low-confident predictions are incorrect.
We assign \textit{random} group labels to those less confident group prediction samples, drawn from the empirical conditional distribution of group labels $a$ given the target labels $y$ (\ie, $P(A|Y=y)$) (Line 4 in \cref{alg:ours}).
In Section \ref{sec:thm}, we make two theoretical contributions. One is to show that our strategy is better than the vanilla pseudo-labeling with respect to the DEO given metric in Section \ref{sec:relwork}, and the other is to show that the random label assignment is equivalent to ignoring the fairness constraint for those random labeled, low-confident samples.
In practice, we expect that our random labeling can play as a regularization method.

For our \methodname, we need one hyperparameter, a confidence threshold $\tau$, to determine whether the given prediction is low confident. We split the given group-labeled training set into training and validation sets (Line 1 and 2 in \cref{alg:ours}) and search the best confidence threshold $\tau$ satisfying the best accuracy on predicting whether the given prediction is correct or wrong (Line 3 in \cref{alg:ours}).
A similar threshold-based strategy is employed in the out-of-distribution sample detection task \cite{hendrycks2016baseline}.
As shown in our experiment, there exists a sweet spot of the confidence threshold $\tau$, where $\tau=1$ is the same as the ``random label'' assignment to all group-unlabeled samples and $\tau=0$ is the same as the vanilla pseudo-labeling strategy.
Once we have the group classifier and the confidence threshold, we assign the pseudo-group labels with our strategy and train the classifier with base off-the-shelf fair-training method on the pseudo-group labeled training samples.
\cref{alg:ours} and \cref{fig:main_figure} illustrate the overview of the proposed \methodname.

\begin{algorithm}[t]
\caption{\methodnamefull}
\label{alg:ours}
\KwData{Group-labeled training set $\mathcal X_L$ and group-unlabeled training set $\mathcal X_U$.}
\KwResult{A set of pseudo group-labels $\tilde A$ for group-unlabeled training set $\mathcal X_U$.}
Split $\mathcal X_L$ into training and validation sets $\mathcal X_L^\text{tr}$, $\mathcal X_L^\text{val}$.\\
Train a group classifier $g:\mathcal{X}\rightarrow S^{|\mathcal{A}|}$ using the training samples $(x, a, y) \sim \mathcal X_L^\text{tr}$, where $S^{|\mathcal{A}|}$ is ${|\mathcal{A}|}$-simplex.\\
Search a confidence threshold $\tau$ on $\mathcal X_L^\text{val}$ that satisfies $\max_{\tau} \sum_{x \in \{x | \max g(x) > \tau\}}\mathbb I(\arg\max g(x) = a) + \sum_{x \in \{x | \max g(x) \leq \tau\}}\mathbb I(\arg\max g(x) \neq a)$.\\
Assign group pseudo-labels $\tilde a$ to $(x, y) \sim \mathcal X_U$ by $\tilde a = \arg\max g(x)$ if $\max g(x) > \tau$, otherwise by sampling from the empirical conditional distribution of $a$ given $y$, \ie, $\tilde a \sim P(A | Y=y)$.
\end{algorithm}
\setlength{\textfloatsep}{4pt}

\subsection{Theoretical understanding of \ours}
\label{sec:thm}
In this subsection, we provide theoretical understandings of why the random label assignment to low confident samples is better than the vanilla pseudo-label (PS) with respect to 
DEO
($\Delta$).
We apply each strategy directly on the true group probability $P(A|X, Y)$, \ie, given an ideally trained group classifier.
Our first theoretical result (Proposition \ref{thm:prop1}) supports that the difference between DEO obtained by our \ours and the underlying DEO is smaller than the difference between the DEO obtained by the vanilla strategy and the underlying DEO.
In other words, our PS with the random assignment is a better approximation of the true $P(A|X, Y)$ than the vanilla PS.
The formal statement is as follows.

\begin{prop}
\label{thm:prop1}
Assume a binary group $\mathcal{A}=\{0,1\}$. 
Let $\Delta(x,y;f,P)$ be the influence of $x$ on DEO, $\Delta(f,P)$ (abbreviated as $\Delta(x,y)$). That is, from $\Delta(f,P) = T(\sum_{x\in \{x|f(x)=1\}} |\Delta(x,y)|)$ where $T(\cdot)$ is the maximum or average over $y$, $\Delta(x,y)$ is defined as follows:
{\small $$\Delta(x,y) \triangleq P(X=x|A=1, Y=y) - P(X=x|A=0,Y=y).$$}%
Let $\widebar P(A|X,Y)$ and $\widehat P(A|X,Y)$ be modified distributions by the vanilla pseudo labeling and \ours, respectively: 
{\small \begin{align}
    &\widebar{P}(A=a|X=x,Y=y) = \mathbb I \left (P(A=a|X=x,Y=y) > 0.5 \right). \nonumber\\
    &\widehat{P}(A=a|X=x,Y=y) \nonumber\\
    &=\begin{cases}
		1, &\text{if } P(A=a|X=x,Y=y) \in [\tau,1]. \\
		P(A=a|Y=y) &\text{if } P(A=a|X=x,Y=y) \in (1-\tau, \tau). \\
		0, &\text{otherwise,}
    \end{cases} \nonumber
\end{align}}%
where $0.5\leq\tau \leq1$ is a threshold value.
Then, we denote $\widebar{P}(X|A,Y)$ and $\widehat{P}(X|A,Y)$ as the distributions induced by $\widebar P(A|X,Y)$ and $\widehat P(A|X,Y)$. We also define $\widebar{\Delta}(x,y)\triangleq{\Delta}(x,y;f,\widebar{P})$ and $\widehat{\Delta}(x,y)\triangleq{\Delta}(x,y;f,\widehat{P})$ as the estimations of $\Delta(x,y)$
Then, for any classifier $f$, each $y$ and all $x \in \{x|f(x)=1 \ \ \text{and}\ \  1-\tau < P(A=1|X=x,Y=y)<\tau \}$, there exists a $\tau$ that satisfies following inequality:
\begin{align}
    |\Delta(x,y)-\widebar{\Delta}(x,y)| > |\Delta(x,y)-\widehat{\Delta}(x,y)| \label{eq:ineq}.
\end{align}
\end{prop}
The proof is in the Appendix.
It helps to get a high-level understanding of the advantages of \ours over the vanilla PS although the Proposition \ref{thm:prop1} is not exactly equivalent to the inequality for $\Delta(f,P)$. 
In practice, due to the simplicity, we approximate the true distribution $P(A=a|X,Y)$ by one group classifier $g$ which learns $P(A=a|X)$, instead of training $|\mathcal{Y}|$ group classifiers for each $y$.
As reported from our experiment in the Appendix,
our group classifier approximates $P(A=a|X)$ well by achieving more than 85\% group accuracy with 50\% of group-labeled training data.

We also show that assigning random labels to a partition of the given data points, $\mathcal X_U$, is equivalent to ignoring the DEO constraint to the data points in $\mathcal X_U$:

\begin{prop}
\label{thm:prop2}
Assume $\mathcal{X}$ is partitioned into any two sets, $\mathcal{X}_L$ and $\mathcal{X}_U$. Let  $\widetilde{P}(A|X,Y)$ be a modified version of $P(A|X,Y)$ as follows:
{\small
\begin{align}
    \widetilde{P}&(A=a|X=x,Y=y) \nonumber\\
    &=\begin{cases}
		P(A=a|X=x,Y=y), &\text{if } x \in \mathcal{X}_L \text{.} \\
		P(A=a|Y=y) &\text{otherwise.}
    \end{cases} \nonumber
\end{align}
}
We denote $\widetilde{P}(X|A,Y)$ as a modified data distribution of $P(X|A,Y)$ induced by $\widetilde{P}(A|X,Y)$.  Then, for any classifier $f$, $\Delta(f,P)$ is the same as $\Delta(f,\widetilde{P})$ using the partial set $\mathcal{X}_L$.
\end{prop}
The proof is in the Appendix.
In practice, since the existing fair-training methods use a relaxed version of DEO,
our method can play as a regularization method to the group-unlabeled samples by assigning random group labels.

\section{Experiments}
In this section, we demonstrate the effectiveness of our \ours for the \ourscenario scenario. We evaluate \ours with various baseline fair-traning methods on three benchmark datasets: UTKFace \cite{utkface} (the sensitive group is ethnicity), CelebA \cite{celeba} (the sensitive group is gender) ProPublica COMPAS \cite{COPMAS} (the sensitive group is ethnicity), and UCI Adult \cite{dua2017uci} (the sensitive group is gender) datasets, where COMPAS and Adult datasets are non-vision tabular datasets. We combine \ours with MFD \cite{jung2021mfd}, FairHSIC \cite{quadrianto2019discovering} and Re-weighting \cite{jiang2020identifying}. To understand the trained group classifier, we provide extensive analysis on the group classifier. Finally, we show the strong empirical contribution of \ours by utilizing extra group-unlabeled training data on the UTKFace dataset. Our \ours shows significant improvements on the target accuracy and the group fairness compared to the baseline methods.

\subsection{Experimental settings}
\subsubsection{Datasets}
\textbf{UTKFace}\cite{utkface}. UTKFace is a facial image dataset, widely adopted as a multi-class and multi-group benchmark. UTKFace contains more than 20K images with annotations, such as age (range from 0 to 116), gender (male and female) and ethnicity (\textit{``White''}, \textit{``Black''}, \textit{``Asian''}, \textit{``Indian''} and \textit{``Others''}). We set ethnicity and age as the sensitive attribute and the target label, respectively. We divided the target age range into three classes: ages between 0 to 19, 20 to 40 and more than 40, following Jung \etal \cite{jung2021mfd}. We use four ethnic groups, \textit{``White''}, \textit{``Black''}, \textit{``Asian''} and \textit{``Indian''}, while \textit{``Others''} is excluded.
The test set is constructed to contain the same number of samples for each group and target.

\noindent\textbf{CelebA}\cite{celeba}. CelebA contains about 200K face images annotated with 40 binary attributes.
As the previous works \cite{jung2021mfd} and \cite{sagawa2019distributionally}, we picked up \textit{``Blond Hair''} and \textit{``Attractive''} as the target labels and \textit{``Gender''}(denoted as \textit{``Male''} in the dataset) as the sensitive attribute.
The results for ``Attractive'' and ethical concerns are provided in the Appendix.
The test set is constructed as the same as UTKFace.

\noindent\textbf{ProPublica COMPAS}\cite{COPMAS}.
We also consider a non-vision tabular dataset to show the versatility of \ours to other modalities.
We use the ProPublica COMPAS dataset, a binary classification task where the target label is whether a defendant reoffends.
We set ethnicity as the sensitive attribute and used the same pre-processing as Bellamy \etal \cite{ai360}, thereby it includes 5,000 data samples, binary group (\textit{``Caucasian''} and \textit{``Non-Caucasian''} and target labels.\vspace{-.1in}\\

\noindent In addition, we also give details and results on \textbf{UCI Adult}\cite{dua2017uci} dataset in the Appendix.

\subsubsection{Base fair-training methods}

We employ three state-of-the-art in-processing methods, \textit{MMD-based Fair Distillation} (MFD) \cite{jung2021mfd}, \textit{FairHSIC} \cite{quadrianto2019discovering} and \textit{Label Bias Correction} (LBC) \cite{jiang2020identifying} for the base fair-training method of \ours.
We briefly describe each method in the 
Appendix.
We only consider scalable fair-training methods for deep learning-based vision applications; note the primal approaches in group fairness \cite{kamishima2012fairness, zafar2017fairness} cannot be applied to the vision domain with high dimensional data and complex models (\eg, DNNs).
We emphasize, however, that our method is not confined to the three methods used in this study but can be easily applied to any fair-training method.

\subsubsection{Implementation details}
\label{subsub:detail}

We provide the implementation details in the Appendix,
including the details of architectures and optimizers, the hyperparameter search protocol. %

\paragraph{Model selection.} \label{paragraph: model selection}
For fairness-aware learning on real datasets, there can exist a trade-off between accuracy and fairness (see an example of the trade-off in \cref{fig:tradeoff}).
For fair comparison, we should select the optimal hyperparmeters showing the best for one criterion while maintaining similar performance for others. Therefore, we select the hyperparameter showing the best fairness criterion $\Delta_M$ while achieving at least 95\% of the vanilla training model accuracy for UTKFace and CelebA datasets. We set the lower bound to 90\% for the COMPAS dataset. If there exists no hyperparameter achieving the minimum target accuracy, we report the hyperparameter with the best accuracy. All models are chosen from the last training epoch.

\begin{figure}[t!]
    \centering
    \begin{subfigure}[b]{\linewidth}
        \centering
        \includegraphics[width=0.9\linewidth]{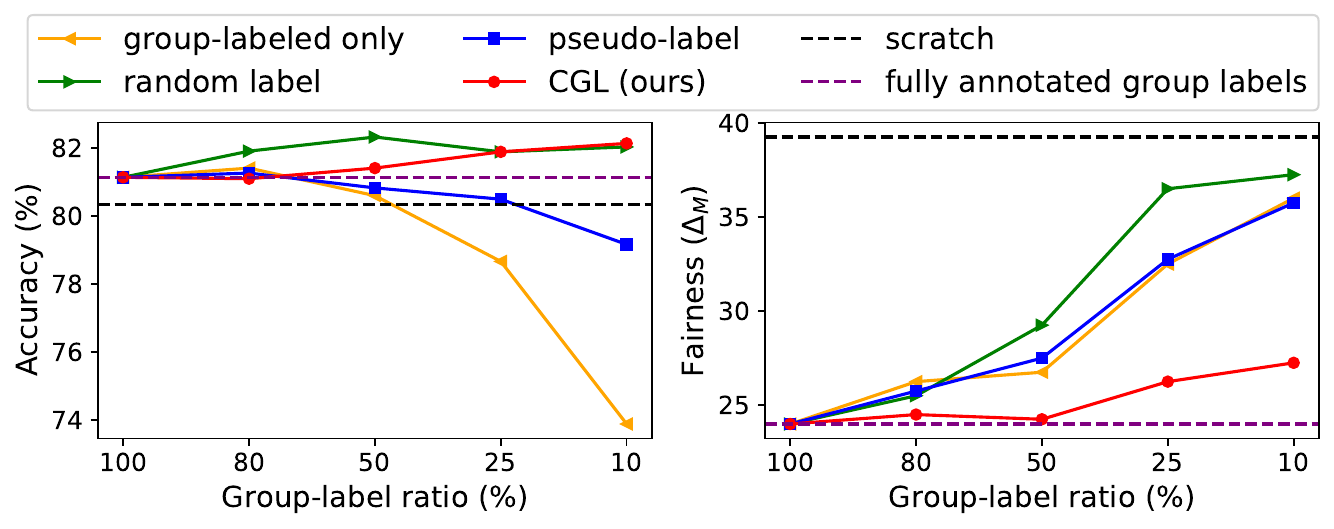}
        \caption{\small MFD results}
    \end{subfigure}
    \begin{subfigure}[b]{\linewidth}
        \centering
        \includegraphics[width=0.9\linewidth]{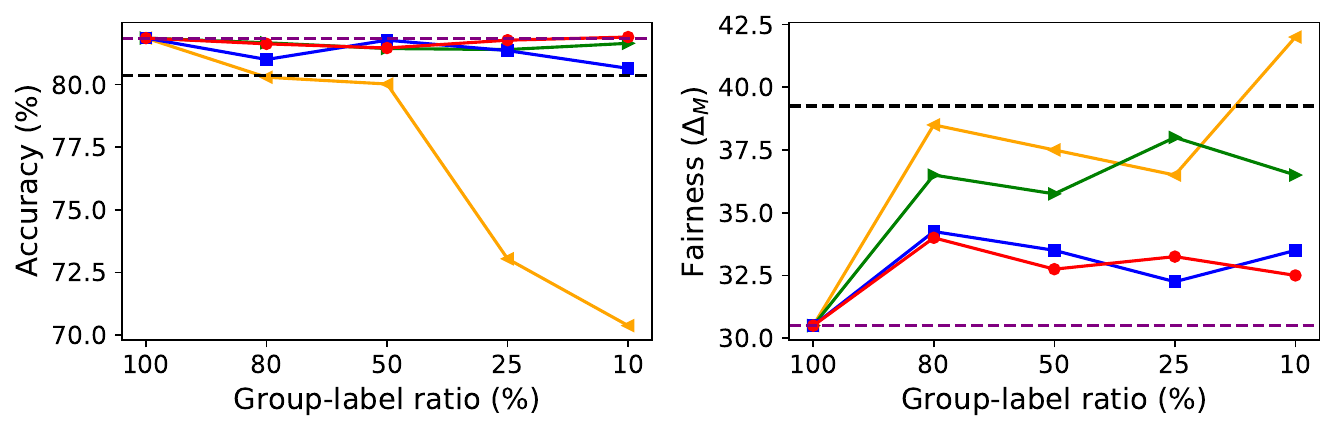}
        \caption{\small FairHSIC results}
    \end{subfigure}
    \begin{subfigure}[b]{\linewidth}
        \centering
        \includegraphics[width=0.9\linewidth]{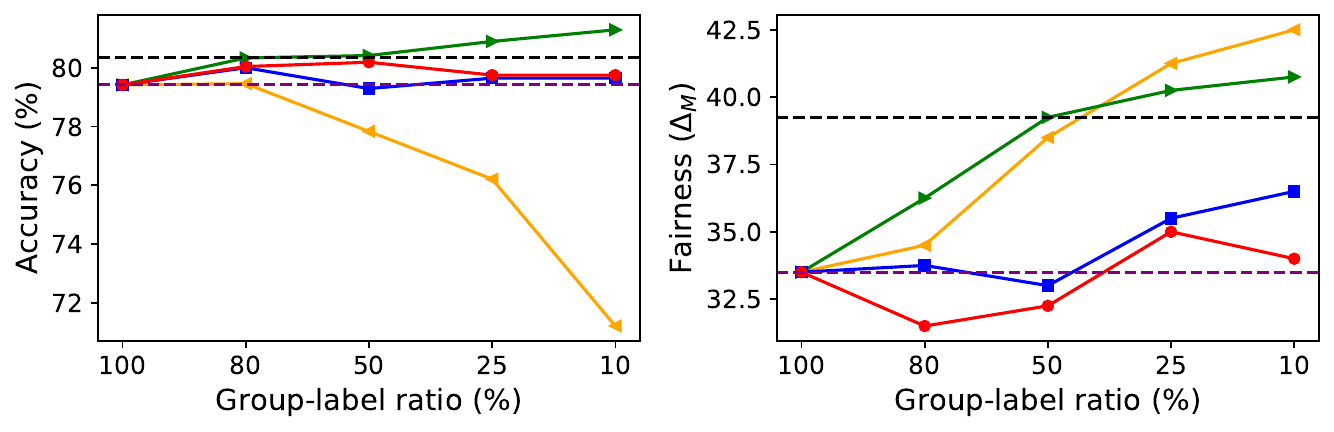}
        \caption{\small LBC results}
    \end{subfigure}
    \caption{\small \textbf{Results on UTKFace.} For varying group-label ratios in training dataset, we show the combination of three fair-training methods with ``group-labeled only'' (yellow), ``random label'' (green), ``pseudo-label'' (blue) and our \ours (red). ``scratch'' denotes the vanilla training without a fairness criteria and ``fully annotated group labels'' denotes the  fair-training methods using the full group labels (\ie, when group-label ratio is 100\%). Higher accuracy and lower $\Delta_M$ denote improvements, respectively.}
    \label{fig:utk_results}
    \vspace{.5em}
\end{figure}

\begin{figure}[t!]
    \centering
    \begin{subfigure}[b]{\linewidth}
        \centering
        \includegraphics[width=0.9\linewidth]{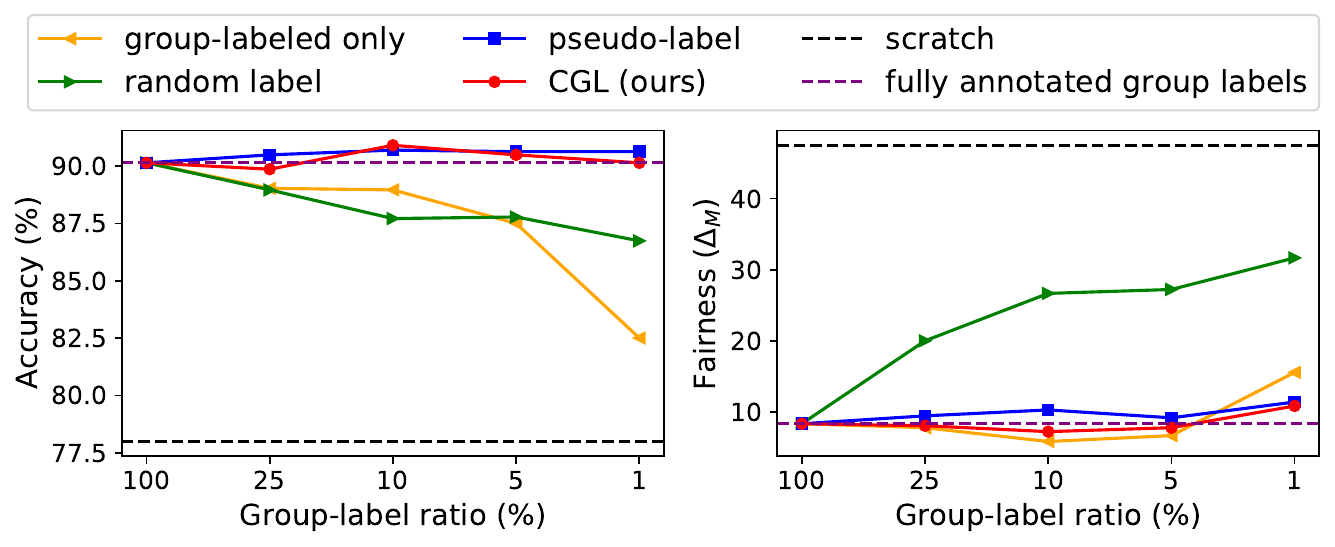}
        \caption{\small MFD results}
    \end{subfigure}
    \begin{subfigure}[b]{\linewidth}
        \centering
        \includegraphics[width=0.9\linewidth]{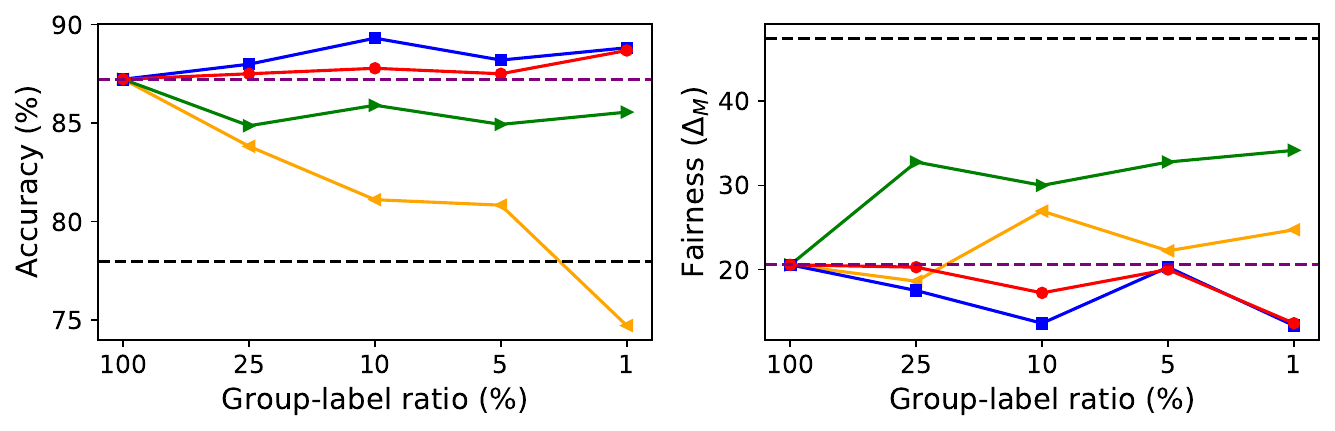}
        \caption{\small FairHSIC results}
    \end{subfigure}
        \begin{subfigure}[b]{\linewidth}
        \centering
        \includegraphics[width=0.9\linewidth]{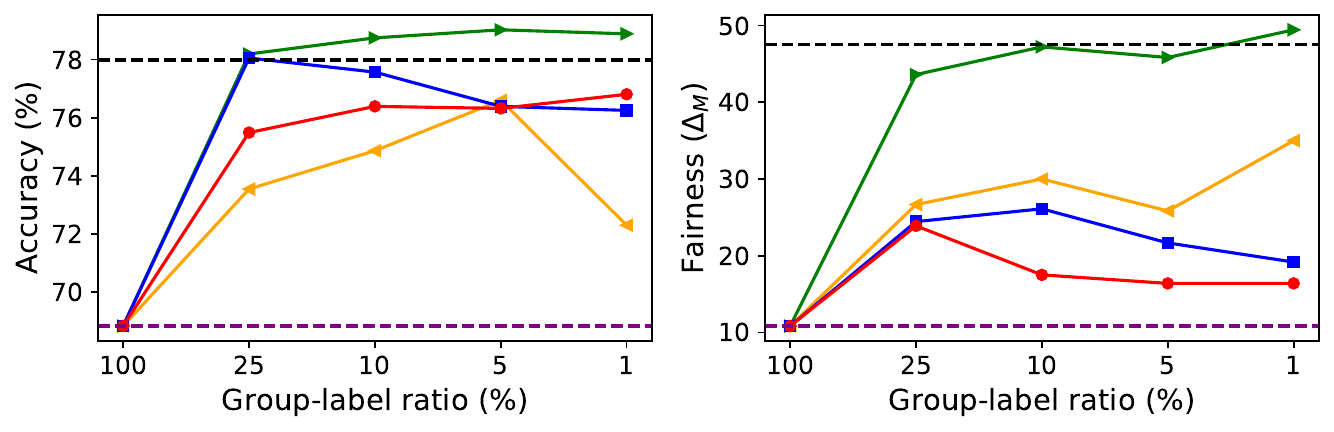}
        \caption{\small LBC results}
    \end{subfigure}
    \caption{\small \textbf{Results on CelebA.} The details are the same as \cref{fig:utk_results}.}
    \label{fig:celeba_results}
    \vspace{.5em}
\end{figure}

\paragraph{Baseline methods and evaluation metrics.}
The existing in-processing methods for group fairness are not applicable directly to our scenario, \ie, when the group labels are not fully annotated. Also, as mentioned in \cref{sec:relwork}, the existing SSL methods mostly cannot be directly applied to \ourscenario as well, since it is not clear whether they achieve the group fairness when applied to predict non-annotated group labels (see the results of UPS \cite{rizve2021ups} in the Appendix, which is one of the state-of-the-art SSL methods). We hence employ three straightforward baselines for comparison.
The \textbf{group-labeled only} strategy discards the group-unlabeled samples and only uses the group-labeled samples for the training. We also examine two group label assignment strategies: The \textbf{random label} strategy assigns random labels to all of the group-unlabeled data (drawn from $P(A|Y=y)$), while the \textbf{pseudo-label} strategy fully trusts the group classifier predictions. Each method is an extreme case of \ours by setting $\tau=1$ and $\tau=0$, respectively. We note that based on the Proposition \ref{thm:prop2}, ``random label'' has the same effect on evaluating the fairness loss part only with the group-labeled samples while evaluating the main loss with all the samples.

We considered three evaluation metrics for all experiments, the target accuracy, $\Delta_M$ and $\Delta_A$ (see \cref{eq:deltam}).
The results are the average scores of four different runs on UTKFace and COMPAS and two different runs on CelebA. $\Delta_A$ and standard deviation scores are given in the Appendix.

\subsection{Main results}

\cref{fig:utk_results} compares the target accuracies and $\Delta_M$ of the combination of MFD, FH and LBC with three baseline strategies
and \ours on the UTKFace dataset with different group-label ratios from 100\% (fully group annotated) to 10 \%.
We show the similar results on the CelebA dataset in \cref{fig:celeba_results} where group-label ratio is chosen from 100\% to 1\%.
Note that we choose different group-label ratios to the datasets because UTKFace is multi-class and multi-group dataset, and CelebA is binary-class and binary-group dataset.
We also emphasize that the performance comparison of the three baselines and \ours mainly focus on $\Delta_M$ because we reported the best $\Delta_M$ of each method among models having accuracy more than the accuracy lower bound described in \cref{paragraph: model selection}. 
\begin{figure}[t!]
    \centering
    \begin{subfigure}[b]{\linewidth}
        \centering
        \includegraphics[width=0.9\linewidth]{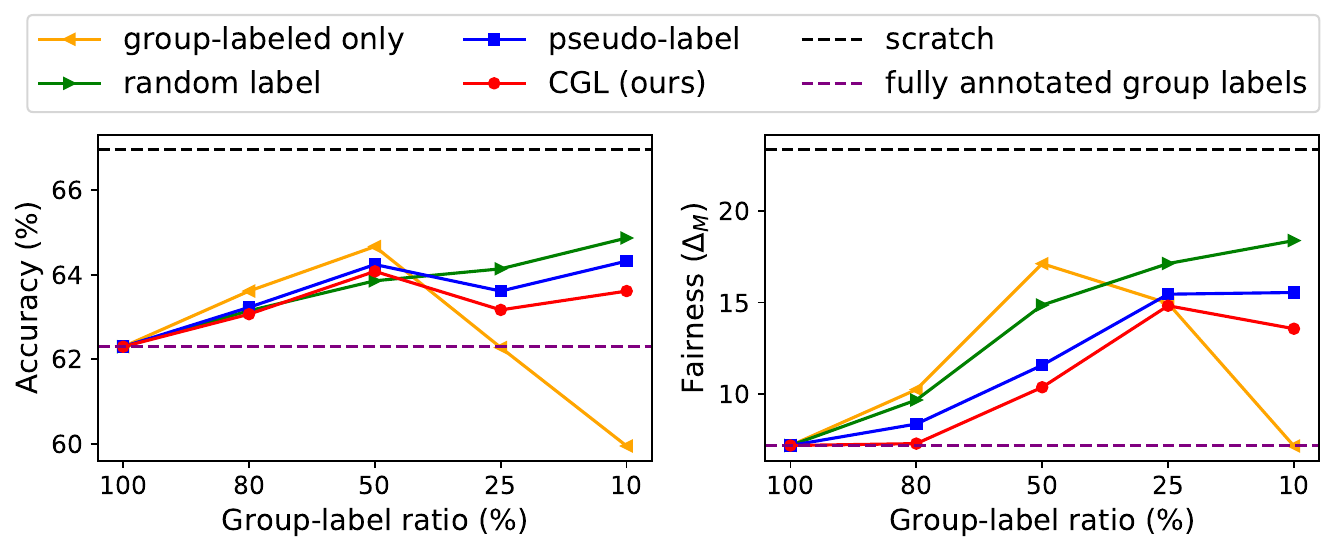}
        \caption{\small MFD results}
    \end{subfigure}
    \begin{subfigure}[b]{\linewidth}
        \centering
        \includegraphics[width=0.9\linewidth]{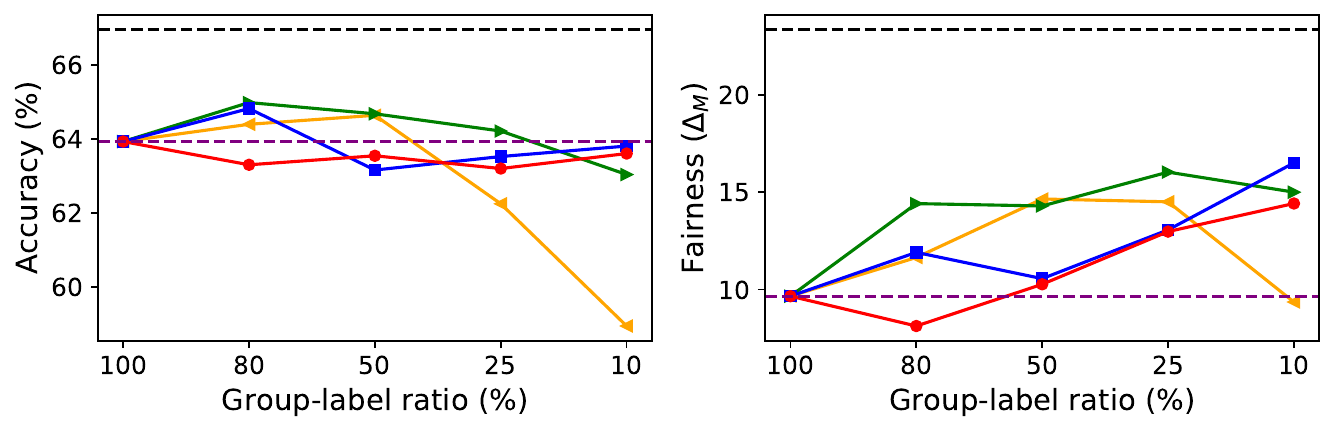}
        \caption{\small FairHSIC results}
    \end{subfigure}
    \begin{subfigure}[b]{\linewidth}
        \centering
        \includegraphics[width=0.9\linewidth]{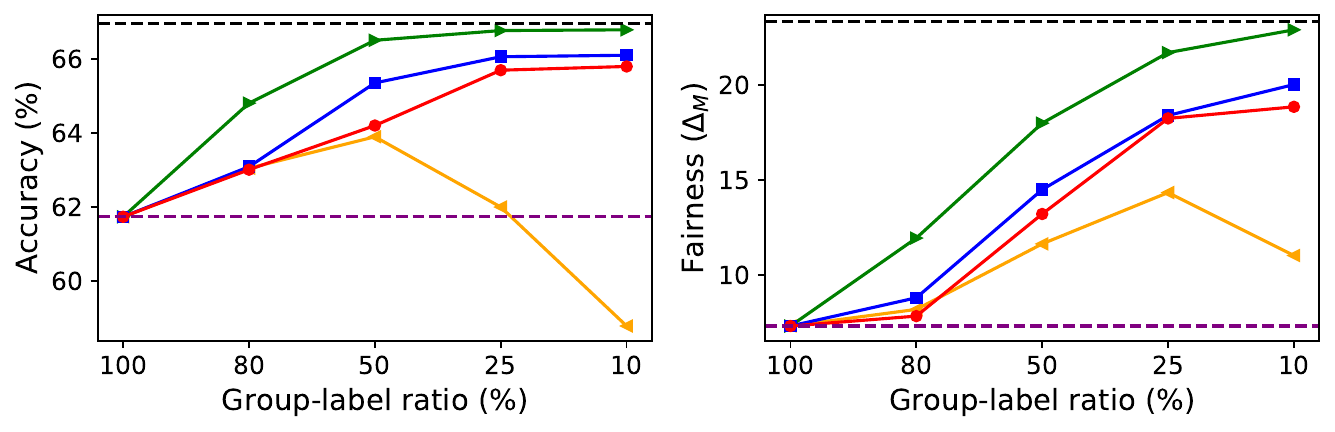}
        \caption{\small LBC results}
    \end{subfigure}
    \caption{\small \textbf{Results on COMPAS.} The details are the same as \cref{fig:utk_results}.}
    \label{fig:compas_results}
\end{figure}

\begin{figure}[t!]
    \centering
    \includegraphics[width=0.52\linewidth]{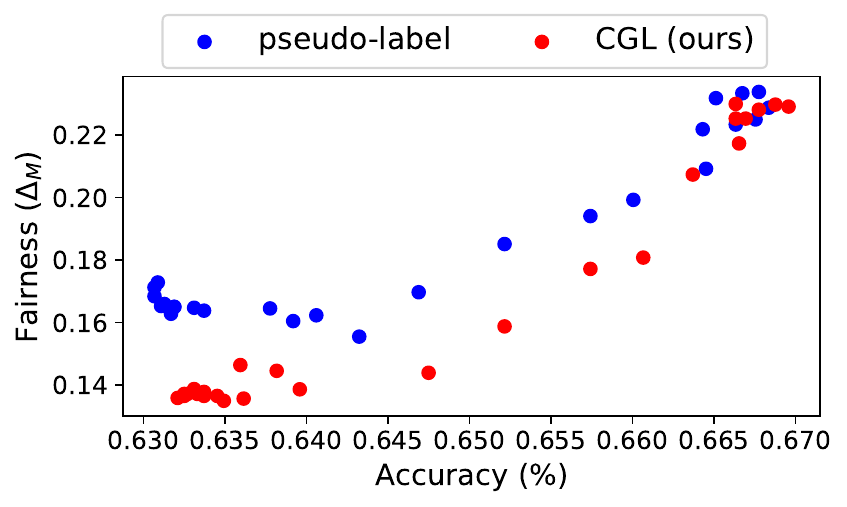}
    \vspace{-.5em}
    \caption{\small {\bf Accuracy-fairness trade-off on COMPAS with 10\% group-labeled training set.} We show accuracy and $\Delta_M$, obtained for CGL and ``pseudo label'' combined with MFD, for different hyperparameters of MFD..}
    \label{fig:tradeoff}
\end{figure}

\begin{figure*}[t!]
    \centering
    \includegraphics[width=0.83\textwidth]{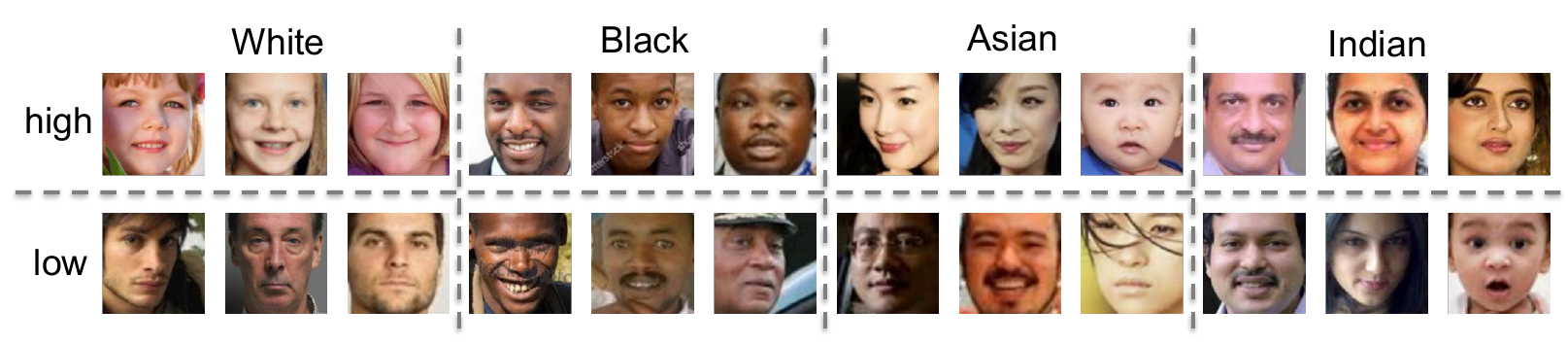}
    \vspace{-1em}
    \caption{\small {\bf High and low confident samples by the group classifier on UTKFace.}We illustrate the top-3 highest and lowest confident samples for that the classifier predicts the correct answer from the UTKFace training samples for each group.}
    \label{fig:visualization}
    \vspace{-1em}
\end{figure*}
In the figures, the ``group-labeled only'' (yellow line) consistently shows much worse accuracies than the baseline methods. Especially, when the group-label ratio decreases, ``group-labeled only'' drastically harms accuracy and fairness at the same time.
The ``random label'' (green line) strategy rarely hurts the accuracies since it uses the full target labels for training, but it shows a drastic drop in $\Delta_M$.
The ``pseudo-label'' (blue line) performs better than the other baselines, but the classifier errors severely affect the fairness performances, especially in the multi-group scenario (\eg, UTKFace).
On the other hand, \ours shows consistently better performances than other baselines in most cases, most notably on UTKFace, by successfully handling samples with low confident group predictions.

We also report the results on non-vision tabular dataset in \cref{fig:compas_results}. We observe similar results to \cref{fig:utk_results} and \cref{fig:celeba_results}.
Note that ``group-labeled only'' shows better fairness criterion in price of the rapid decrease of accuracy in the low group label regime.
Our method generally performs better than other baselines in all methods in terms of fairness. We point out that although the accuracies of \ours are slightly lower than those of the other baselines, it does not necessarily mean that those schemes perform better since they must sacrifice much more accuracy to achieve similar $\Delta_M$ to \ours.
To clarify this, we plotted the full accuracy-fairness trade-offs with different hyperparameters on COMPAS in \cref{fig:tradeoff}. We clearly observe that CGL dominates ``pseudo label'' by achieving a better Pareto trade-off curve, which implies the validity of our model selection rule given in Section \ref{paragraph: model selection}.

\subsection{Analysis of group classifiers}
\paragraph{Group classifier confidences.}
We show the highest and lowest confident samples by the group classifier on UTKFace in \cref{fig:visualization}.
As shown in the figure, the low confident samples are qualitatively uncertain to humans due to diverse lighting, various orientations and low quality. Therefore, our confidence-based thresholding can capture the inherent uncertainty of the dataset. 
In the Appendix,
we provide the confidence score distribution and the group classifier accuracies for different group label ratios.

\begin{figure}[t!]
    \centering
    \includegraphics[width=0.43\textwidth]{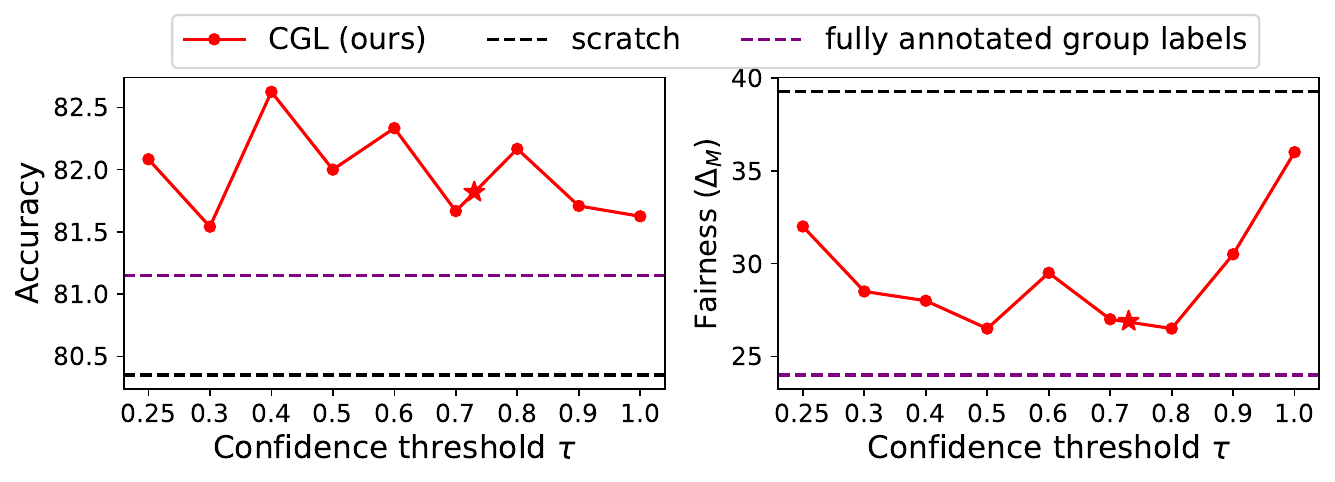}
    \vspace{-.5em}
    \caption{\small {\bf $\tau$ study on UTKFace.} Accuracies and fairness (by $\Delta_M$) for varying  $\tau$. $\tau\leq0.25$ is the same as ``random label'' and $\tau=1$ corresponds to vanilla ``pseudo-label'' in other figures.} 
    \label{fig:tau_study}
    \vspace{.5em}
\end{figure}

\paragraph{Study on the threshold $\tau$.}
Figure \ref{fig:tau_study} shows the accuracies and $\Delta_M$ of \ours and MFD by varying $\tau$ on UTKFace with 10\% group-labeled training set.
We fix the hyperparameters used in \cref{fig:utk_results} and report the average of two different runs.
$\tau\leq0.25$ (since there are four groups) and $\tau=1$ is equivalent to ``random label'' and ``pseudo-label'' in the previous results, respectively. The ``$\star$''  represents the results of $\tau$ achieved by the our strategy (Line 4 in \cref{alg:ours}).
Here, we observe that there exists a sweet spot of the threshold that shows better $\Delta_M$ and accuracy than ``random label'' and ``pseudo-label'', and our method can achieve a good threshold near the sweet spot. 

\subsection{Augmenting with extra group-unlabeled data}
\label{sec:extradata}

We finally show the impact of the \ourscenario scenario and our \ours on the UTKFace dataset and an extra group-unlabeled dataset.
We use the FairFace dataset \cite{karkkainen2021fairface} for the extra dataset. FairFace contains 108,501 facial images with balanced attributes. We filter out ethnicity not in ``White'', ``Black'', ``Asian'' and ``Indian''. After the filtering, we have 73,377 extra samples.
To examine our \ourscenario problem, we let the extra datasets only have the target labels (\ie, ages) but not the group labels.
\cref{tab:extradata} shows the results of the scratch and MFD trained only on the UTKFace, and the scratch and MFD+\ours on the UTKFace augmented with the FairFace dataset as above.
Interestingly, MFD on UTKFace only shows worse fairness ($\Delta_M=25.0$) than the scratch training on the UTKFace + FairFace ($\Delta_M=24.0$), which is in line with the result on low group label regime in (\cref{fig:teaser}).
We achieve the state-of-the-art accuracy (84.38\%) and fairness ($\Delta_M=19.5$) by successfully augmenting UTKFace with the extra group-unlabeled dataset.

\begin{table}[t]
    \centering
    \small
    \caption{\small {\bf Impact of \ours on UTKFace and extra group-unlabeled training dataset.} The accuracy and fairness criterion on the UTKFace test set are shown.
    For ``MFD + CGL'', we assign group psuedo-labels by \ours to the extra group-unlabeled samples from FairFace (73,377 images) with the group classidier trained on full UTKFace training set (20,813 images). We then
    train MFD on the psuedo-labeled training dataset (94,190 images).}
    \label{tab:extradata}
    \resizebox{\columnwidth}{!} {
    \begin{tabular}{ccccc}
    \toprule
    & \multicolumn{2}{c}{UTKFace only} & \multicolumn{2}{c}{UTKFace + FairFace}\\
    & Scratch & MFD & Scratch & MFD + \ours \\ \midrule
Accuracy ($\uparrow$)  & 80.29      & 83.46    & 81.15       & \textbf{84.38}     \\
Fairness $\Delta_A$ ($\downarrow$) & 20.17      & 16.67    & 15.67       & \textbf{13.00}     \\
Fairness $\Delta_M$ ($\downarrow$) & 39.00      & 25.00    & 24.00       & \textbf{19.50} \\ \bottomrule
    \end{tabular}
    }
\end{table}

\section{Concluding Remark}

We considered a practical learning scenario in which the group labels are partially annotated for fariness-aware learning.
We have observed that the existing fair-training method can even perform worse than the scratch training when the number of group labels is small.
We propose a simple yet effective solution that is readily applicable to any fair-training method and demonstrated that \ours improves various baselines on several benchmarks.
We believe our method can significantly reduce the cost for obtaining additional group labels for all the training samples, facilitating more rapid development of fair classifiers.

\section*{Acknowledgments}

This work was supported in part by the New Faculty Startup Fund from Seoul National University, NRF
Mid-Career Research Program [NRF-2021R1A2C2007884], IITP grants funded by the Korean government [No.2019-0-01396, Development of framework
for analyzing, detecting, mitigating of bias in AI model and training data], [No.2021-0-01343, Artificial Intelligence Graduate School Program (Seoul National University)], [No.2021-0-02068, Artificial Intelligence
Innovation Hub (Artificial Intelligence Institute, Seoul National University)], and SNU-NAVER Hyperscale AI Center.

{\small
\bibliographystyle{ieee_fullname}
\bibliography{egbib}
}

\clearpage
\onecolumn
\appendix
\numberwithin{equation}{section}
\numberwithin{figure}{section}
\numberwithin{table}{section}
\section*{Supplementary Materials}
We include additional materials in this document. We first state our societal impact, dataset license, limitations and ethical concerns in the beginning. We provide additional related works for biases in machine learning in \cref{sec: more  works} and a detailed proof of our propositions in \cref{appendix-sec:proof}. We include our implementation details, such as architecture, optimization, hyperparameter search and base fairness methods and their modifications in \cref{appendix-sec:implementation}. We provide the additional analysis of group classifiers in \cref{appendix-sec:groupclsf}, experimental results in \cref{appendix-sec:more results} and result tables in \cref{appendix-sec:result tabels}.

\noindent\paragraph{Dataset license.}
In the paper, we use four datasets: UTKFace \cite{utkface}, CelebA \cite{celeba}, ProPublica COMPAS \cite{COPMAS} and FairFace \cite{karkkainen2021fairface}.
According to the official web page\footnote{\url{https://susanqq.github.io/UTKFace/}}, UTKFace dataset is a non-commercial license dataset where the copyright belongs to the original owners in the web. The dataset is built by Dlib \cite{dlib09} and annotations are tagged by the DEX algorithm and human annotators.
CelebA dataset has a similar license statement\footnote{\url{https://mmlab.ie.cuhk.edu.hk/projects/CelebA.html}} to UTKFace.
COMPAS dataset is collected its data points from Broward County Sheriff's Office in Florida\footnote{\url{https://www.propublica.org/article/how-we-analyzed-the-compas-recidivism-algorithm}} which is a public records.
FairFace is licensed by CC by 4.0\footnote{\url{https://github.com/joojs/fairface}}.
Overall, all datasets have clean licenses that is applicable to any public research project.

\paragraph{Societal impact.}
As we stated in the main text, a vanilla DNN training can occur negative societal impacts by dismissing fairness criterion, on the other hand, considering fairness criterion at the training time requires a huge number of group labels. We expect our \ours can bridge the gap between real-world applications and fairness-aware training, so that mitigating the negative societal impacts economically by only annotating a subset of group-unlabeled samples.

\noindent\paragraph{Limitations.}
Although our method can be applied to any fairness method, we observe that \ours is not always better than other baselines.
First, our method relies on the quality of group classifier, hence, if the group classifier performs worse, our method does not guarantee better fairness than the vanilla pseudo-labeling.
Also, the group classifier predictions can be noisy. In Appendix, we show group prediction accuracy of our group classifier. In the low group label regime, the accuracy of our classifier decreases to less than 80\% on UTKFace. This implies that if the base method is sensitive to noisy group labels (\eg, Adversairal De-biasing), our method and pseudo-labeling can perform worse than our expectation.
Finally, in the case that a distribution shift for the sensitive attribute exists when predicting group labels of group-unlabeled data from group-unlabeled data, the naive application of would suffer from performance degradation. These distribution shift can be alleviated by training a group classifier with robust optimization techniques (\eg, choosing a distribution shift-aware optimizer \cite{cha2021swad}, invariant risk minimization \cite{arjovsky2019invariant} or group distributed robust optimization \cite{sagawa2019distributionally}). 

\noindent \paragraph{Ethical concerns}
We originally used a subjective and potentially unethical ``Attractive'' attribute in our experiments with the CelebA dataset. It is known that ``Attractive'' is highly correlated to gender (``Male''), while most other attributes are not \cite{torfason2016face}. Our purpose of CelebA experiments is to show the scalability of our method as CelebA (200K) is a large-scale dataset compared to UTK (20K), COMPAS (5K), Adult (40K). From a similar motivation, many previous studies employed Attractive as their target label \cite{jung2021mfd, chuang2020fair, park2021learning, quadrianto2019discovering}. Particularly, Quadrianto et al. used Attractive ``as the proxy measure of getting invited for a job interview in the world of fame'' \cite{quadrianto2019discovering}. However, we agree that using a subjective attribute as ``Attractive'' can be unethical. We only used the results as an example, and we alert that such classifiers for attractiveness can cause potential ethical concerns.

\section{Proof of propositions}
\label{appendix-sec:proof}
\subsection{Proof of Proposition 1}
\begin{proof}
We only show only the case where $P(A=1|X=x,Y=y)\geq0.5$ and the opposite case can be proved in the same way. For any classifier $f$ and all $x\in \{x|f(x)=1 \, \text{and } 0.5\leq P(A=1|X=x, Y=y)<\tau\}$, we have from $\widebar{P}$ and $\widehat{P}$ defined in (Eq. (3) and (4), manuscript),
\begin{align}
\widebar{\Delta}(x,y) &= \big(\frac{1}{P(A=1|Y=y)}\big)P(X=x|Y=y), \\
\widehat{\Delta}(x,y) &= 0.
\end{align}
Then, we have
\begin{align}
\label{eq:case}
&|\Delta(x,y) - \widebar{\Delta}(x,y)| - |\Delta(x,y) - \widehat{\Delta}(x,y)| 
= \begin{cases}
\widebar{\Delta}(x,y) \quad &\text{if } \Delta(x,y)\leq 0 \\
\widebar{\Delta}(x,y) - 2\Delta(x,y) \quad &\text{otherwise.}
\end{cases}
\end{align}
For the first case in \cref{eq:case}, we can trivially see that $\Delta(x,y)>0$. For the second case in \cref{eq:case}, we have
\begin{align}
&\widebar{\Delta}(x,y) - 2\Delta(x,y) \\
&= \big(\frac{1-2P(A=1|X=x,Y=y)}{P(A=1|Y=y)} + \frac{2P(A=0|X=x,Y=y)}{P(A=0|Y=y)}\big)P(X=x|Y=y) > 0
\end{align}
, if $P(A=1|X=x,Y=y) < \frac{P(A=1|Y=y)+1}{2}$. Therefore, we have the proposition 1 by setting $\tau$ to $\frac{P(A=1|Y=y)+1}{2}$.
\end{proof}

\subsection{Proof of Proposition 2}
\begin{proof}
Given a data distribution $P(X,A,Y)$ and a classifier $f$, $\Delta(f,P)$ is defined as follows:
\begin{align}
    \label{eq: def_delta}
   & \Delta(f, P) =T\bigg(\max_{a,a'} \big(| \underbrace{P(\hat{Y}=y|A=a, Y=y) - P(\hat{Y}=y|A=a',Y=y)}_{(a)} | \big) \bigg)
\end{align}
, where $T(\cdot)$ can be the maximum or average over $y$ depending on the types of $\Delta$.
For each $y, a$ and $a'$, the above argument of $\max_{a,a'}$, (a) in \cref{eq: def_delta}, can be represented as follows:
\begin{align}
(a) &= \sum_{x\in \{x|f(x)=y\}} P(X=x|A=a,Y=y)-P(X=x|A=a',Y=y) \nonumber\\
&= \sum_{x\in \{x\in X_L|f(x)=y\}} P(X=x|A=a,Y=y)-P(X=x|A=a',Y=y) \nonumber \\
    & \quad + \sum_{x\in \{x\in X_U|f(x)=1\}} P(X=x|A=a,Y=y)-P(X=x|A=a',Y=y) \label{eq: xl_xu}
\end{align}
Then, the second term of \cref{eq: xl_xu} can be represented as follows:
\begin{align}
     &\sum_{x\in \{x\in X_U|f(x)=y\}} P(X=x|A=a,Y=y)-P(X=x|A=a',Y=y) \nonumber\\  
     & = \sum_{x\in \{x\in X_U|f(x)=y\}} \frac{P(A=a|X=x,Y=y)P(X=x|Y=y)}{P(A=a|Y=y)} - \frac{P(A=a'|X=x,Y=y)P(X=x|Y=y)}{P(A=a'|Y=y)}\label{eq:c3} 
\end{align}
If we substitute $P(A|X,Y)$ into $\widehat{P}(A|X,Y)$ in the RHS of \cref{eq:c3}, we have the proposition 2.
\end{proof}

\section{Additional Related Works for Biases in Machine Learning}
\label{sec: more works}
Emerging studies on DNNs have revealed that DNNs rely on shortcut biases \cite{cadene2019rubi, geirhos2019imagenet, bahng2019rebias, geirhos2020shortcut, scimeca2021wcst-ml}.
The existing de-biasing methods let a model less attend on the dataset biases in an implicit way by using extra biased networks \cite{cadene2019rubi, bahng2019rebias} or data augmentations \cite{geirhos2019imagenet} without using bias labels. Both fairness methods and de-biasing methods aim to learn a representation invariant to undesired decision cues, such as sensitive groups and dataset biases. However, de-biasing methods explore implicit shortcut biases that harm the network generalizability, where many known shortcuts (\eg, language bias \cite{cadene2019rubi} or texture bias \cite{geirhos2019imagenet}) are neither strongly relative to ethical concerns nor easy to configure. On the other hand, in the fairness problem, sensitive groups are diversely defined by the target application to avoid negative societal impacts (\ie, a model should make the same predictions to any social group
such as ethnicity or gender). Therefore, even though de-biasing methods can be applied to \ourscenario by ignoring group labels, there is no guarantee to learn fair models by the de-biasing approaches. In this work, we focus on fairness methods explicitly utilizing group labels for the base method of \ours. 

\section{More Implementation Details}
\label{appendix-sec:implementation}

\subsection{Architecture and optimization}
\label{appendix-sec:implementation-arch}
We choose the same architecture for the base classifier and the group classifier; ResNet18 \cite{he2016deep} for the UTKFace and CelebA experiments and a simple 2-layered neural network for the COMPAS experiments. On UKTFace and CelebA datasets, we train the models with the Adam optimizer \cite{kingma2014adam} for 70 epochs by setting the initial learning rate 0.001 reduced by 0.1 when the loss is stagnated for 10 epochs following Jung \etal \cite{jung2021mfd}. We train the model for 50 epochs on COMPAS dataset.
All results are reported by the model at the last epoch.

\subsection{Hyperparameter search}
In the experiments, there are two types of hyperparameters: the confidence threshold of \ours, and the method-specific hyperparameters for each method.
Since our method only needs the group-labeled training dataset for training group classifier and seeking a threshold, we split the group-labeled samples into 80\% training and 20\% validation samples. The confidence threshold is searched on the validation set (by Algorithm 1, manuscript).

Fairness-aware training methods are usually sensitive to the hyperparameter selection due to the accuracy-fairness trade-off; when the strength for fairness is getting stronger, the target accuracy is getting worse. For example, a trivial solution to achieve the fairest classifier is to predict all labels to a constant label, while this solution is the worst solution in terms of the target accuracy. Hence, the careful tuning of the control parameters to fairness criteria (\eg, MMD \cite{jung2021mfd}, HSIC \cite{quadrianto2019discovering} or adversarial loss \cite{adv_debiasing}) takes the key role in handling the accuracy-fairness trade-off.
In our experiments, we aim to find a fair classifier while showing \textit{a comparable accuracy} to the vanilla training method. Thus, we select the hyperparameter showing the best fairness criterion $\Delta_M$ while achieving at least 95\% of the vanilla training model accuracy. We set the lower bound to 90\% for the COPMAS dataset. If there exists no hyperparameter achieving the minimum target accuracy, we report the hyperparameter with the best accuracy.
We perform the grid search on the hyperparameter candidates for every partial group-label case and for every method. The full hyperparameter search space is illustrated in \cref{appendix-tab:hyperparams}.

\begin{table}[t]
    \centering
    \begin{tabular}{ccc}
    \toprule
        Method & Hyperparameter & Candidates \\ \midrule
        MFD \cite{jung2021mfd} & MMD strength $\lambda$ &  [10, 30, 100, 300, 1000, 3000, 10000, 30000] \\ \midrule
        FairHSIC \cite{quadrianto2019discovering} & HSIC strength $\lambda$ & [1,3,10, 30, 100, 300, 1000, 3000] \\ \midrule
        \multirow{2}{*}{LBC \cite{jiang2020identifying}}  & Adversary strength $\alpha$ & [1, 3, 10, 30, 100] \\
                                                          & learning rate of adversary & [$10^{-4}$, $10^{-2}$] \\ \bottomrule
    \end{tabular}
    \caption{\small \textbf{Hyperparameter search spaces.} We perform the grid search on the validation set to find the best hyperparameters for each method. We use the same hyperparameters for optimizer (See \cref{appendix-sec:implementation-arch}).}
    \label{appendix-tab:hyperparams}
\end{table}

\subsection{Base fairness methods and their modifications}
\label{appendix-sec:implementation-basemethods}

Here, we describe the overview of each base fairness method used for the experiments.
MFD and FairHSIC use additional fairness-aware regularization terms as the relaxed version of the targeted fairness criteria.
MFD proposed a \textit{maximum-mean-discrepancy}-based \cite{gretton2012kernel} regularization term to achieve fairness via feature distillation and FairHSIC devised a \textit{HSIC}-based \cite{gretton2005measuring} regularization term to obtain feature representations independent on group labels.
For FairHSIC, we only implement the second term of their decomposition loss (\ie, the HSIC loss between the feature representations and the group labels).

LBC is a re-weighting algorithm optimizing weights of examples through multiple iterations of full training to ensure their theoretical guarantees.
The original LBD requires multiple full-training iterations by alternatively computing a EO criterion after full-training and re-training the full dataset by optimal weights. This alternative optimization needs a very huge training budget. We modify the EO computation iteration to a few-epoch iterations, \ie, 5 epochs, instead of the full-training.

AD lets an adversary cannot predict group labels by the additional adversarial loss. In our experiments, AD shows little improvements if the group or target label is not binary where Jung \etal \cite{jung2021mfd} witnessed the same phenomenon.
Thus, we use multiple adversaries for AD to make AD be available to solve multi-class and multi-group problems following Jung \etal \cite{jung2021mfd} and omit the loss projection in the original objectives of AD for a stable learning.
Also, we only report AD results for the Compas dataset while AD does not perform well on other vision datasets.

\section{Additional Analysis of Group Classifiers}
\label{appendix-sec:groupclsf}

\begin{figure}[t!]
    \centering
    \includegraphics[width=0.4\linewidth]{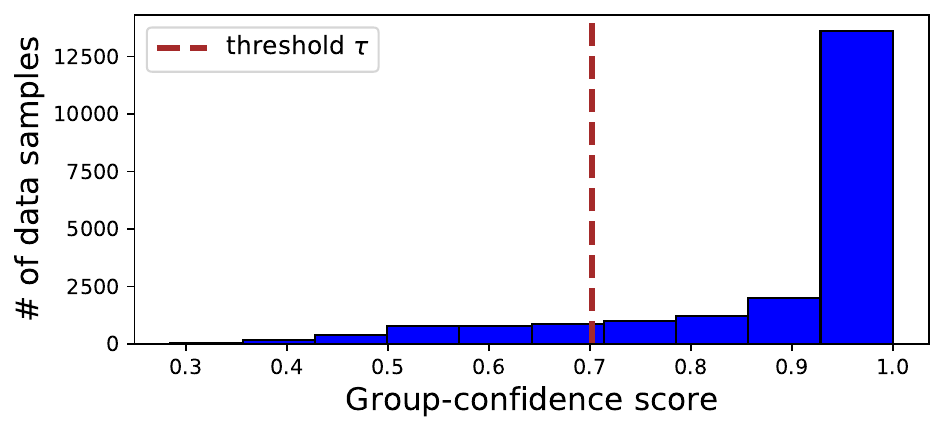}
    \caption{\small {\bf Group confidences verse sample densities.} The number of samples for each confidence bin is shown. The red dotted line denotes the selected threshold in the UTKFace experiments.}
    \label{fig:confidence_dist}
\end{figure}

\begin{table}[t!]
\small
\centering
\caption{\small \textbf{Group classifier performances.} We compare the accuracies by the baseline decision rule ($\arg\max$) and by our method (assigning random labels to low confident samples) for the trained group classifiers on the small group-labeled training samples.}
\label{tab:groupclassifierperf}
\begin{tabular}{ccccc}
\toprule
Group-label ratio & 80\%   & 50\%   & 25\%  & 10\%   \\ \midrule
Baseline & 87.88 & 86.11 & 82.82 & 77.73 \\ 
Ours & 87.24 & 85.81 & 82.59 & 75.21\\ \bottomrule
\end{tabular}
\end{table}

\begin{figure}[t!]
    \centering
    \includegraphics[width=0.6\linewidth]{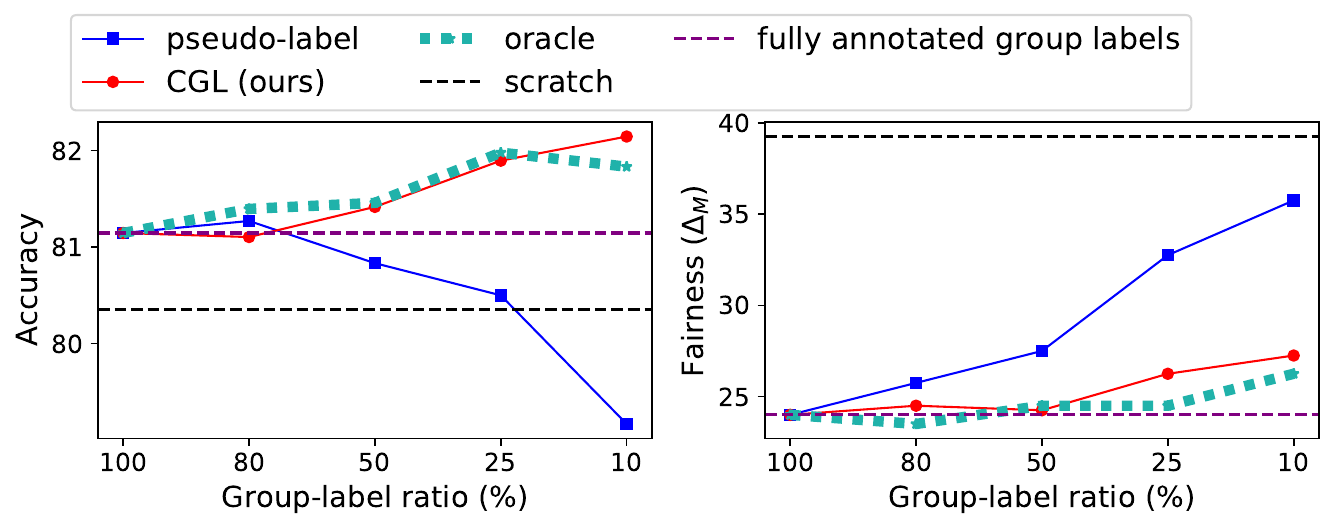}
    \caption{\small \textbf{Comparisons with an ``oracle'' fair group classifier on UTKFace with MFD.} The oracle classifier group classifier has the same accuracy with our group classifier (used for ``pseudo-label'' and ``\ours (ours)'' -- See \cref{tab:groupclassifierperf}) but the wrong samples by the ``oracle'' classifier are \textit{randomly} chosen from the dataset.}
    \label{fig:oracle_results}
\end{figure}

\paragraph{Prediction confidences by our group classifier.}
In the main manuscript, we show the highest and lowest confident samples by the group classifier on UTKFace in 
Fig. 7.
As shown in the figure, low confident samples are qualitatively uncertain to humans due to diverse lighting, various orientations and low quality, where Shi \etal observed the same results by an uncertainty-aware face embedding \cite{shi2019probabilistic}.
From the qualitative results, we observe that our confidence-based threshold method can reasonably capture the inherent uncertainty of the dataset without an explicit uncertainty-aware training, such as MC-Dropout \cite{gal2016dropout} or probabilistic embeddings \cite{oh2019iclr, chun2021pcme}.

However, because our group classifier does not guarantee to capture proper uncertainty measures, we presume that applying an uncertainty-aware training can improve \ours as Rizve \etal \cite{rizve2021ups}.
We show the number of samples by the confidences in \cref{fig:confidence_dist}.
Our classifier shows high confident predictions (over 65\% predictions are confident than 0.9 because) because it is not trained by calibration-aware regularizations \cite{guo2017calibration} or other regularization techniques known to help confidence calibration scores \cite{chun2019icmlw}, such as mixed sample augmentations \cite{zhang2017mixup, yun2019cutmix} and smoothed labels \cite{labelsmoothing}.
Nonetheless, we observe that many images are still low confident and our group classifier can figure whether the prediction is correct or wrong; when we apply the optimal threshold, our classifier has 85.43\% accuracy to figure out whether the prediction is wrong or correct.

\paragraph{Quality of our group classifier and the threshold-based decision rule.}
In \cref{tab:groupclassifierperf}, we show the group accuracies of our group classifier by different decision rules on varying group label ratio. We show two different decision rules: the baseline $\arg\max$ strategy and our confidence-based random altering (\ie, $\arg\max$ if the confidence is larger than $\tau$, otherwise $P(A|Y)$) with the best threshold. We observe that our random label strategy slightly hurts the accuracies but not significantly. In other words, our group classifier has well-sorted confidences that can capture the self predictive uncertainty.

Finally, we compare our group classifier and the ``oracle'' group classifier which has the same accuracy to ours, but group labels that our group classifier wrongly predict are replaced into a group label sampled from an uniform distribution. In other words, ``oracle'' assumes the scenario where our confidence-based thresholding perfectly operates. 
\cref{fig:oracle_results} shows the comparison of \ours, ``pseudo-label'' and ``oracle'' on UTKFace dataset and MFD. Here, we see that ``oracle'' significantly improve the performance in terms of fairness other than ``pseudo-label''. This imply that only random-labeling for wrongly predicted group labels can prevent performance degradation of DEO, which experimentally supports our proposition 2. We also observe that the performance of \ours is comparable one of ``oracle'', meaning that random labeling low confident samples are more critical to the performance than high confident samples with noisy group labels.

\section{Additional experimental results}
\label{appendix-sec:more results}

\begin{figure}[t!]
    \centering
    \begin{subfigure}{0.48\textwidth}
        \includegraphics[width=\linewidth]{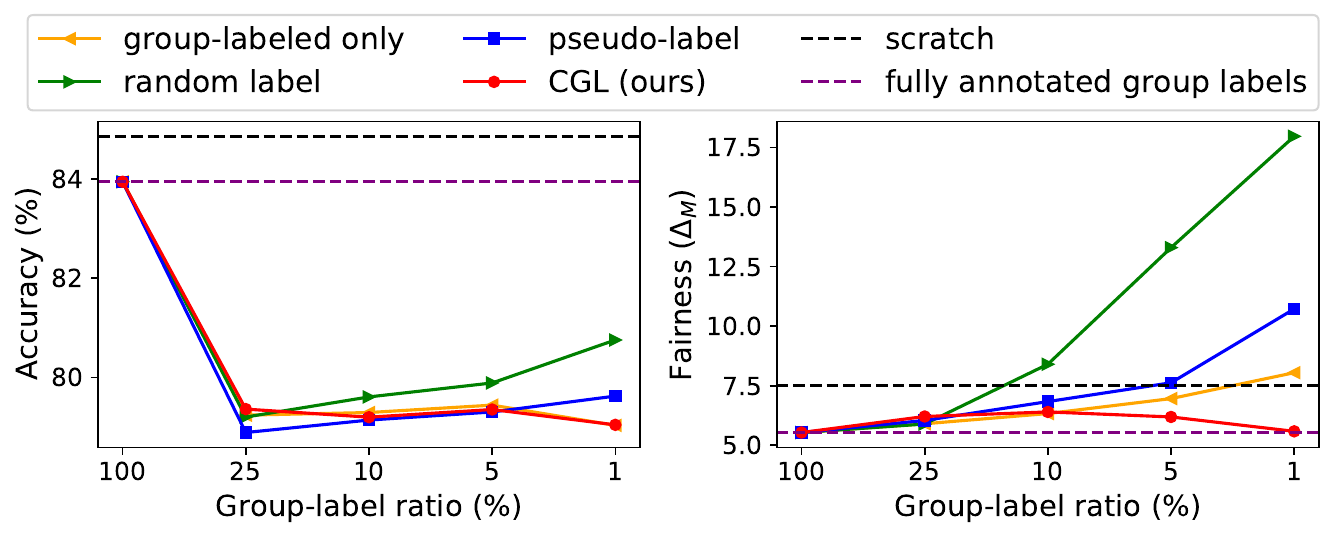}
        \caption{\small MFD results on Adult}
    \end{subfigure} \hspace*{\fill}
    \begin{subfigure}{0.48\textwidth}
        \includegraphics[width=\linewidth]{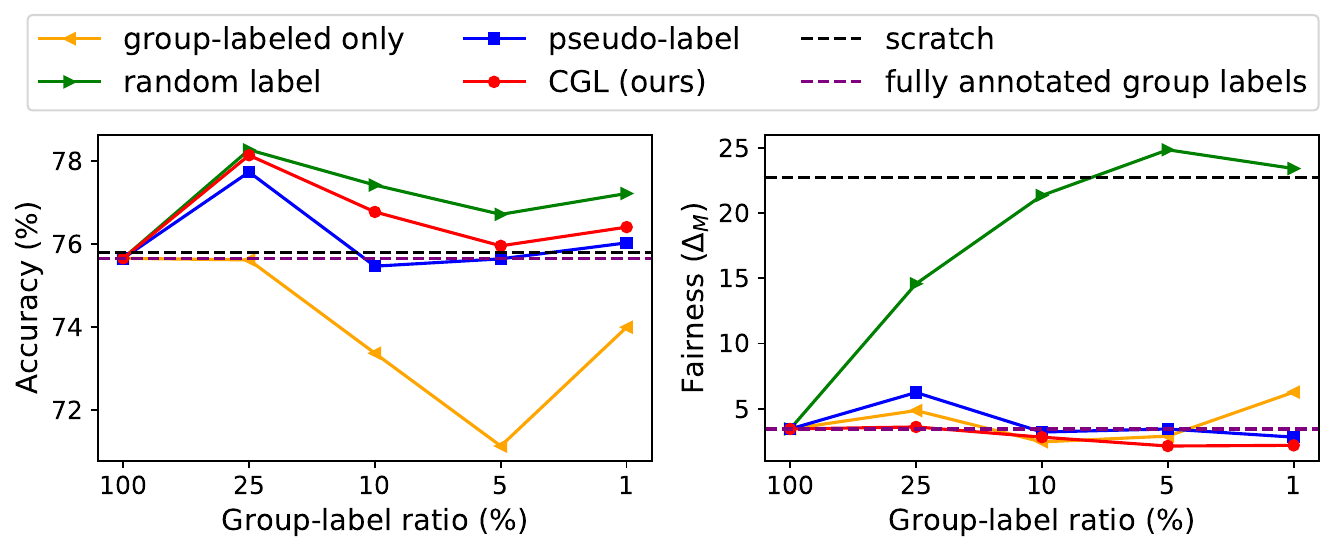}
        \caption{\small MFD results on CelebA}
    \end{subfigure}    
    
    \begin{subfigure}{0.48\textwidth}
        \includegraphics[width=\linewidth]{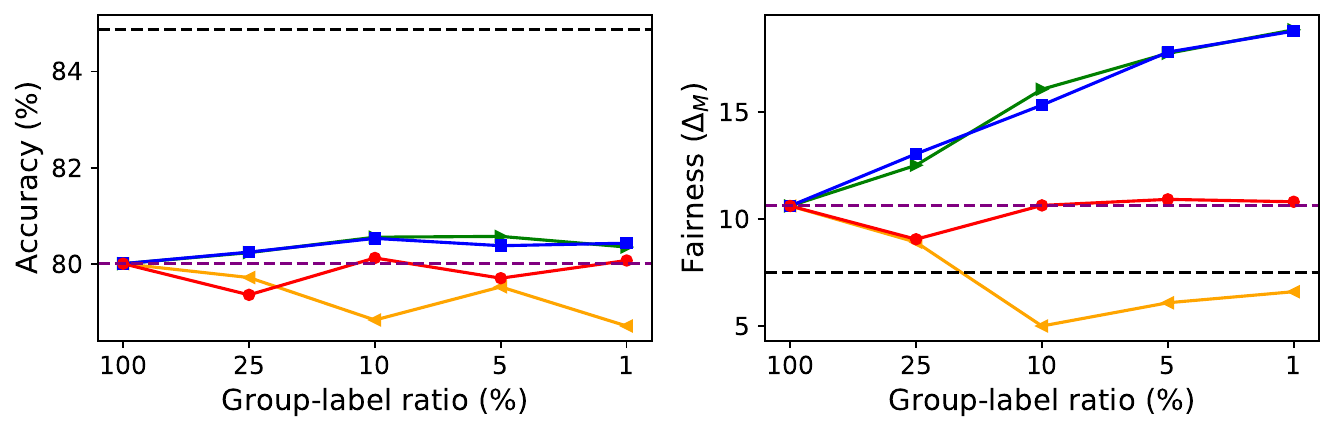}
        \caption{\small FairHSIC results on Adult}
    \end{subfigure} \hspace*{\fill}    
    \begin{subfigure}{0.48\textwidth}
        \includegraphics[width=\linewidth]{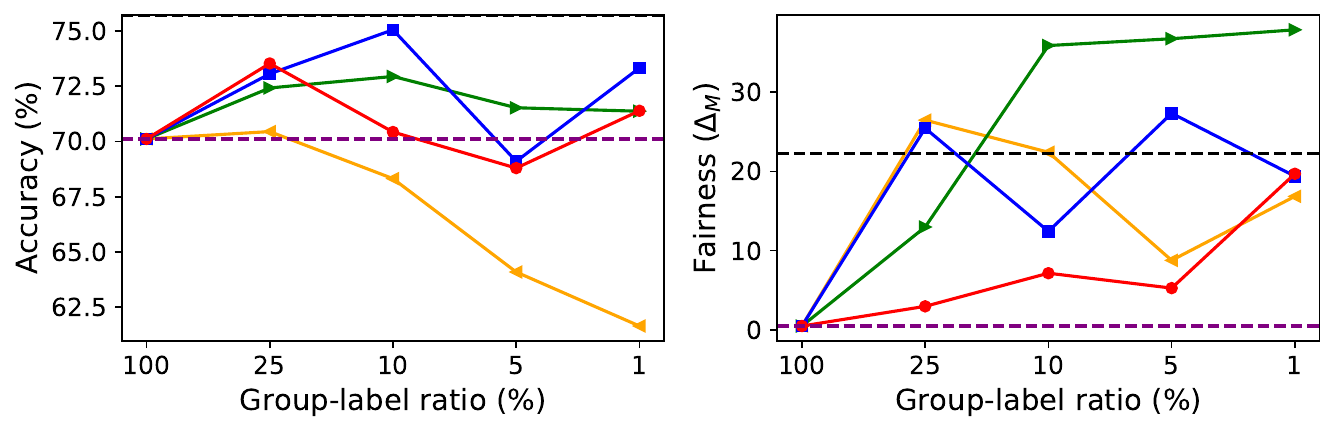}
        \caption{\small FairHSIC results on CelebA}
    \end{subfigure}
    
    \begin{subfigure}{0.48\textwidth}
        \includegraphics[width=\linewidth]{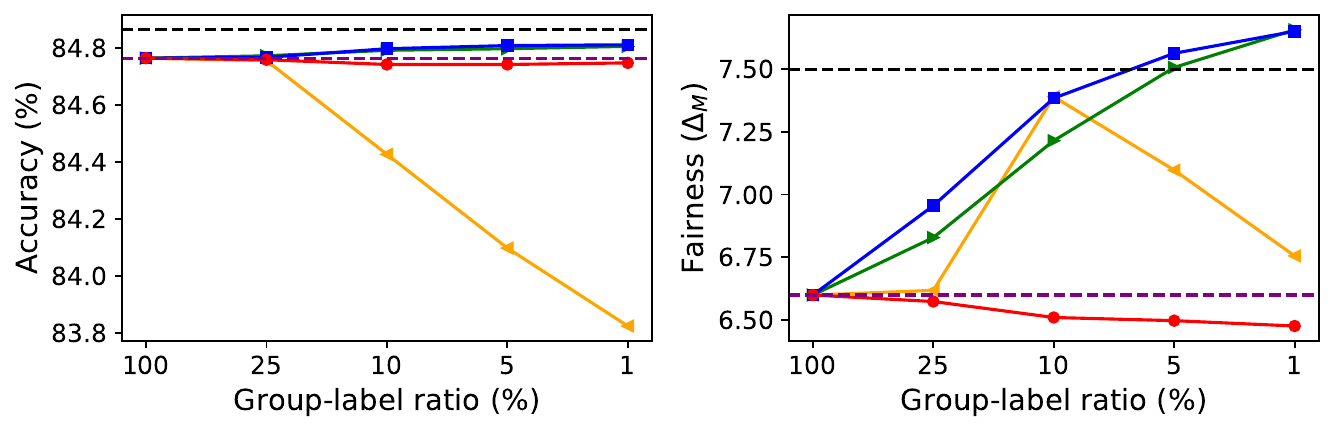}
        \caption{\small LBC results on Adult}
    \end{subfigure} \hspace*{\fill}    
    \begin{subfigure}{0.48\textwidth}
        \includegraphics[width=\linewidth]{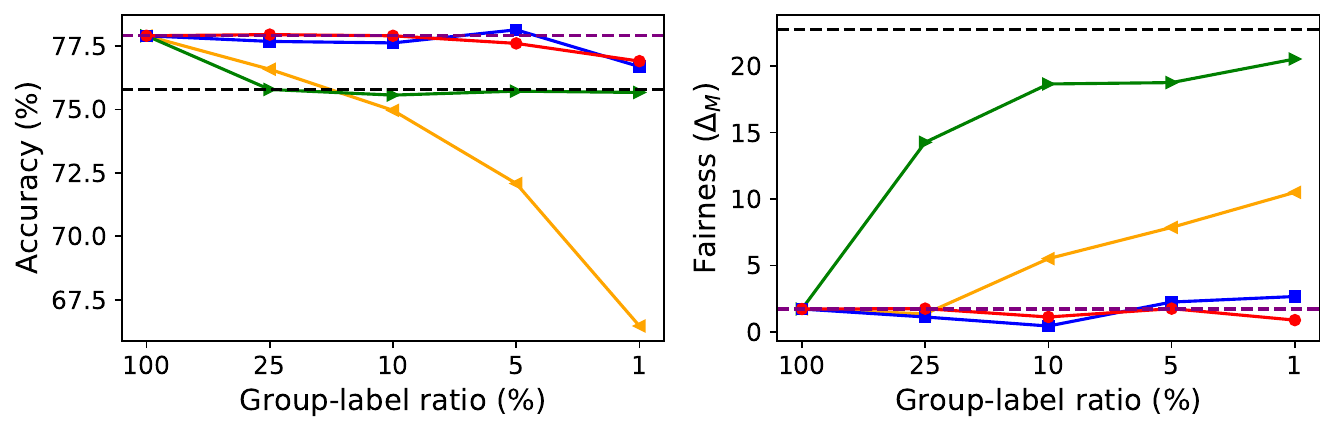}
        \caption{\small LBC results on CelebA}
    \end{subfigure}
    \caption{\small \textbf{Results on Adult and CelebA. The target label in CelebA is ``Attractive'' attribute.} The details are the same as Fig. 3.
    }
    \label{fig:adult_celeba_results attractive}
\end{figure}

\subsection{Results on Adult dataset}
\label{subsec: adult results}
To show the consistent improvements on another dataset, we conducted an additional experiment on Adult dataset with the same details as the main experiment in the manuscript. UCI Adult dataset \cite{dua2017uci} is a non-vision tabular dataset used for a binary classification task where the target label is whether the income exceeds \$50K per a year given attributes about the person. We set gender as the sensitive attribute and used the same processing as Bellamy \etal. \cite{ai360}, so that it includes 45,000 data samples. 

The left column of \cref{fig:adult_celeba_results attractive} shows the results of CGL and baselines combined with base fairness methods on Adult dataset, and we observe the consistent trend of CGL that our method mostly performs better than other baselines for fairness. We repeatedly note that our slightly lower accuracies do not imply the ineffectiveness of CGL because we report the model with the best DEO where accuracy is lower-bounded.

\subsection{Results on CelebA using the ``Attractive'' attribute as the target label}
\label{subsec: celeba attractive results}

The right column of \cref{fig:adult_celeba_results attractive} shows the target accuracy and $\Delta_M$ on CelebA using ``Attractive'' attribute as the target label. From the right column of 
\cref{fig:adult_celeba_results attractive},
we again demonstrate the better performance of CGL than other baselines for all base fairness methods. Since the ``Attractive'' attribute would be the subjective and potentially unethical to discuss the results rigorously, as described in the beginning of Appendix, we advise that these results should used only as a auxiliary and not as a primary result.

\subsection{Comparison CGL with UPS}
\label{subsec: UPS results}
\begin{figure}[t!]
    \centering
    \includegraphics[width=0.6\linewidth]{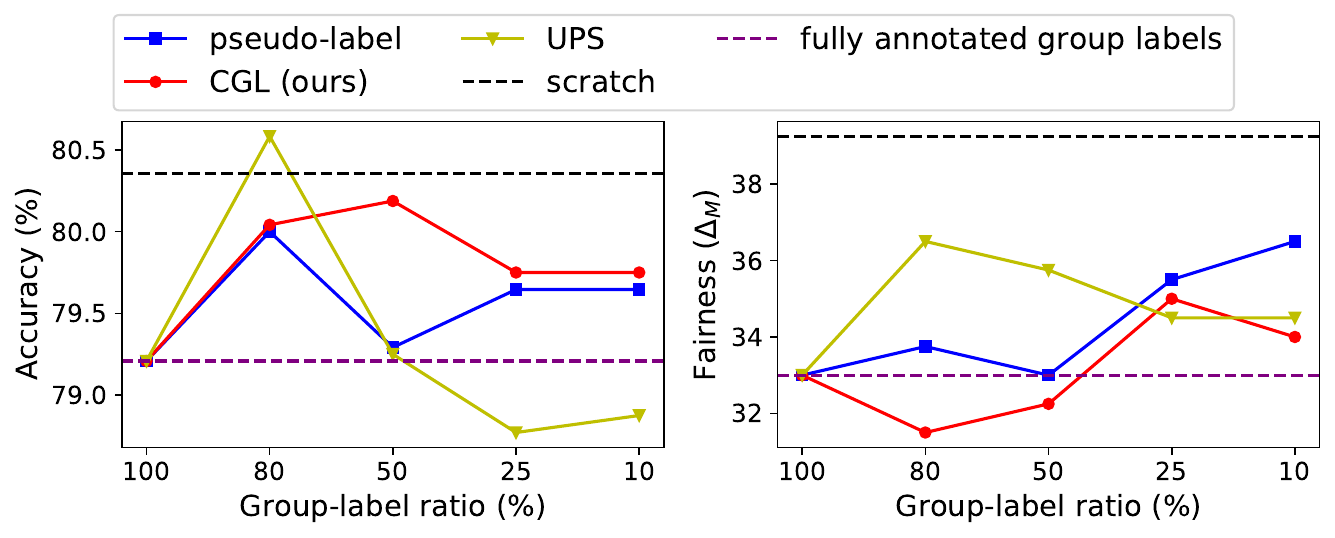}
    \caption{LBC results on UTKFace}
    \label{fig:ups results}
\end{figure}
The aim for SSL is to simply predict the future attribute labels as accurately as possible from the partial annotations in the training set, it is not clear whether the predicted attribute labels can be directly plugged-in to achieve the group fairness in the test set. To corroborate our finding, we carried out additional experiments with a state-of-the-art SSL method, UPS, utilized for Fair-PG. UPS iteratively trains the group classifier and predicts the missing group labels in the training set and filters out the samples with uncertain predictions. (We omitted the negative learning of UPS since it cannot be applied any base fairness methods.) Note such filtering would unnecessarily discard significant amount of the target label information, hence, the accuracy would hurt particularly when the group label ratio is low. In \cref{fig:ups results}, we report the result of LBC on UTKFace, including the UPS baseline. We indeed observe that UPS suffers from low accuracy especially when the group-label ratio is low, and CGL mostly outperforms UPS for both accuracy and fairness. This confirms that a naive plug-in of SSL method for Fair-PG would not be satisfactory.

\subsection{AD results on COMPAS dataset}
\label{subsec:ad results}

\begin{table}[t]
\centering
\caption{\small \textbf{Accuracy on COPMAS for AD.}}
\label{tab:acc-Co-A}
\begin{tabular}{cccccc}
\toprule
                  & 100\%             & 80\%              & 50\%              & 25\%              & 10\%              \\ \midrule
group-labeled only & \multirow{4}{*}{63.51 ($\pm$1.45)} & 65.32 ($\pm$0.58)&63.65 ($\pm$0.37)&61.30 ($\pm$1.22)&57.52 ($\pm$2.84) \\
random label       &  & 63.61 ($\pm$0.55)&63.11 ($\pm$0.67)&64.44 ($\pm$1.38)&64.67 ($\pm$0.24) \\
psuedo-label       &  & 64.55 ($\pm$0.41)&64.12 ($\pm$0.63)&63.19 ($\pm$0.18)&65.80 ($\pm$0.38) \\
CGL                &  & 63.05 ($\pm$1.13)&63.25 ($\pm$0.60)&64.24 ($\pm$1.24)&63.82 ($\pm$1.55) \\ \bottomrule
\end{tabular}
\end{table}

\begin{table}[t]
\centering
\caption{\small \textbf{$\Delta_A$ on COPMAS for AD.}}
\label{tab:deltaA-Co-A}
\begin{tabular}{cccccc}
\toprule
                  & 100\%             & 80\%              & 50\%              & 25\%              & 10\%              \\ \midrule
group-labeled only & \multirow{4}{*}{10.35 ($\pm$1.84)} & 13.32 ($\pm$2.14)&11.46 ($\pm$0.63)&9.75 ($\pm$1.84)&5.27 ($\pm$0.76) \\
random label       &  & 9.26 ($\pm$1.46)&10.69 ($\pm$1.46)&13.17 ($\pm$2.10)&11.90 ($\pm$1.44) \\
psuedo-label       &  &12.43 ($\pm$3.39)&12.11 ($\pm$4.07)&11.37 ($\pm$3.17)&16.26 ($\pm$0.57) \\
CGL                &  & 9.63 ($\pm$3.60)&11.93 ($\pm$3.90)&14.71 ($\pm$1.27)&10.67 ($\pm$2.70) \\ \bottomrule
\end{tabular}
\end{table}

\begin{table}[t]
\centering
\caption{\small \textbf{$\Delta_M$ on COPMAS for AD.}}
\label{tab:deltaM-Co-A}
\begin{tabular}{cccccc}
\toprule
                  & 100\%             & 80\%              & 50\%              & 25\%              & 10\%              \\ \midrule
group-labeled only & \multirow{4}{*}{12.72 ($\pm$2.98)} & 16.30 ($\pm$2.41)&14.39 ($\pm$1.50)&12.61 ($\pm$2.11)&8.52 ($\pm$2.22) \\
random label       &  & 12.37 ($\pm$2.09)&13.51 ($\pm$1.38)&15.96 ($\pm$1.93)&15.70 ($\pm$2.26) \\
psuedo-label       &  & 16.15 ($\pm$3.79)&15.68 ($\pm$4.73)&13.97 ($\pm$2.67)&19.57 ($\pm$0.93) \\
CGL                &  & 13.78 ($\pm$5.00)&14.73 ($\pm$5.28)&17.96 ($\pm$0.31)&13.23 ($\pm$3.82) \\ \bottomrule
\end{tabular}
\end{table}
\cref{tab:acc-Co-A}, \cref{tab:deltaA-Co-A} and \cref{tab:deltaM-Co-A} compare the target accuraies, $\Delta_A$ and $\Delta_M$ of the combinations of AD with three baselines and CGL on COMPAS dataset. The number in the parentheses with $\pm$ stands for the standard deviation of each metric obtained several independent runs with different seeds. our CGL again show the better performances than other baselines in terms of fairness for most cases. Through the case where the group-label ratio is 25\%, we can see that confidence-based thresholding by a group classifier can be slightly sensitive in the group label regime if the base fairness method is vulnerable to noisy group labels (\eg, AD).

\subsection{Result tables}
\label{appendix-sec:result tabels}
Table from \ref{tab:acc-U-M} to \ref{tab:deltaM-Co-L} show the detailed results including accuracy, $\Delta_A$ and $\Delta_M$ for all experiments in Figure 3, 4 and 5 in the main manuscript.
The details of numbers in parentheses are the same as tables in \cref{subsec:ad results}. 

\begin{table}[t]
\centering
\caption{\small \textbf{Accuracy on UTKFace for MFD.}}
\label{tab:acc-U-M}
\begin{tabular}{cccccc}
\toprule
                  & 100\%             & 80\%              & 50\%              & 25\%              & 10\%              \\ \midrule
group-labeled only & \multirow{4}{*}{81.15 ($\pm$0.28)} & 81.42 ($\pm$0.39)&80.60 ($\pm$0.37)&78.67 ($\pm$0.64)&73.88 ($\pm$0.78) \\
random label       &  & 81.92 ($\pm$0.36)&82.33 ($\pm$0.53)&81.90 ($\pm$0.63)&82.04 ($\pm$0.34) \\
psuedo-label       &  & 81.27 ($\pm$0.60)&80.83 ($\pm$0.39)&80.50 ($\pm$0.54)&79.17 ($\pm$0.54) \\
CGL                &  & 81.10 ($\pm$0.24)&81.42 ($\pm$0.42)&81.90 ($\pm$0.41)&82.15 ($\pm$0.58) \\ \bottomrule
\end{tabular}
\end{table}

\begin{table}[t]
\centering
\caption{\small \textbf{$\Delta_A$ on UTKFace for MFD.}}
\begin{tabular}{cccccc}
\toprule
                  & 100\%             & 80\%              & 50\%              & 25\%              & 10\%              \\ \midrule
group-labeled only & \multirow{4}{*}{15.67 ($\pm$0.71)} & 16.33 ($\pm$0.85)&17.08 ($\pm$1.46)&18.50 ($\pm$1.38)&21.25 ($\pm$2.66) \\
random label       &  & 16.83 ($\pm$0.29)&18.58 ($\pm$0.83)&22.58 ($\pm$0.86)&23.50 ($\pm$1.80) \\
psuedo-label       &  & 16.33 ($\pm$0.97)&16.67 ($\pm$0.41)&18.58 ($\pm$1.95)&20.00 ($\pm$2.16) \\
CGL                &  & 15.33 ($\pm$1.03)&14.92 ($\pm$2.17)&17.17 ($\pm$1.57)&17.25 ($\pm$1.04) \\ \bottomrule
\end{tabular}
\end{table}

\begin{table}[t]
\centering
\caption{\small \textbf{$\Delta_M$ on UTKFace for MFD.}}
\begin{tabular}{cccccc}
\toprule
                  & 100\%             & 80\%              & 50\%              & 25\%              & 10\%              \\ \midrule
group-labeled only & \multirow{4}{*}{24.00 ($\pm$1.58)} & 26.25 ($\pm$3.56)&26.75 ($\pm$2.59)&32.50 ($\pm$2.87)&36.00 ($\pm$2.92) \\
random label       &  & 25.50 ($\pm$1.66)&29.25 ($\pm$4.66)&36.50 ($\pm$0.50)&37.25 ($\pm$3.19) \\
psuedo-label       &  & 25.75 ($\pm$2.86)&27.50 ($\pm$0.87)&32.75 ($\pm$3.83)&35.75 ($\pm$4.49) \\
CGL                &  & 24.50 ($\pm$2.06)&24.25 ($\pm$2.17)&26.25 ($\pm$3.49)&27.25 ($\pm$2.77) \\ \bottomrule
\end{tabular}
\end{table}

\begin{table}[t]
\centering
\caption{\small \textbf{Accuracy on UTKFace for FairHSIC.}}
\begin{tabular}{cccccc}
\toprule
                  & 100\%             & 80\%              & 50\%              & 25\%              & 10\%              \\ \midrule
group-labeled only & \multirow{4}{*}{81.85 ($\pm$0.23)} & 80.29 ($\pm$0.64)&80.02 ($\pm$1.10)&73.04 ($\pm$3.68)&70.38 ($\pm$1.27) \\
random label       &  & 81.67 ($\pm$0.48)&81.44 ($\pm$0.78)&81.40 ($\pm$0.78)&81.65 ($\pm$0.56) \\
psuedo-label       &  & 81.00 ($\pm$1.02)&81.77 ($\pm$0.26)&81.35 ($\pm$0.56)&80.65 ($\pm$0.59) \\
CGL                &  & 81.62 ($\pm$0.79)&81.46 ($\pm$0.72)&81.77 ($\pm$0.57)&81.90 ($\pm$0.89) \\ \bottomrule
\end{tabular}
\end{table}

\begin{table}[t]
\centering
\caption{\small \textbf{$\Delta_A$ on UTKFace for FairHSIC.}}
\begin{tabular}{cccccc}
\toprule
                  & 100\%             & 80\%              & 50\%              & 25\%              & 10\%              \\ \midrule
group-labeled only & \multirow{4}{*}{18.50 ($\pm$1.67)} & 21.33 ($\pm$1.62)&21.67 ($\pm$1.67)&22.08 ($\pm$2.18)&27.42 ($\pm$4.30) \\
random label       &  & 22.50 ($\pm$1.71)&22.50 ($\pm$1.30)&23.75 ($\pm$2.17)&23.50 ($\pm$1.34) \\
psuedo-label       &  & 21.92 ($\pm$1.01)&21.08 ($\pm$2.25)&19.75 ($\pm$1.77)&20.67 ($\pm$0.94) \\
CGL                &  & 20.67 ($\pm$1.70)&20.75 ($\pm$1.09)&20.42 ($\pm$1.11)&18.50 ($\pm$1.46) \\ \bottomrule
\end{tabular}
\end{table}

\begin{table}[t]
\centering
\caption{\small \textbf{$\Delta_M$ on UTKFace for FairHSIC.}}
\begin{tabular}{cccccc}
\toprule
                  & 100\%             & 80\%              & 50\%              & 25\%              & 10\%              \\ \midrule
group-labeled only & \multirow{4}{*}{30.50 ($\pm$4.33)} & 38.50 ($\pm$2.96)&37.50 ($\pm$3.84)&36.50 ($\pm$2.18)&42.00 ($\pm$3.67) \\
random label       &  & 36.50 ($\pm$3.04)&35.75 ($\pm$3.27)&38.00 ($\pm$3.67)&36.50 ($\pm$2.60) \\
psuedo-label       &  & 34.25 ($\pm$3.27)&33.50 ($\pm$1.50)&32.25 ($\pm$4.97)&33.50 ($\pm$1.66) \\
CGL                &  & 34.00 ($\pm$3.08)&32.75 ($\pm$2.28)&33.25 ($\pm$2.86)&32.50 ($\pm$2.69) \\ \bottomrule
\end{tabular}
\end{table}

\begin{table}[t]
\centering
\caption{\small \textbf{Accuracy on UTKFace for LBC.}}
\begin{tabular}{cccccc}
\toprule
                  & 100\%             & 80\%              & 50\%              & 25\%              & 10\%              \\ \midrule
group-labeled only & \multirow{4}{*}{79.42 ($\pm$0.74)} & 79.46 ($\pm$1.16)&77.83 ($\pm$0.28)&76.21 ($\pm$0.63)&71.21 ($\pm$1.06) \\
random label       &  & 80.33 ($\pm$0.69)&80.42 ($\pm$0.64)&80.90 ($\pm$0.62)&81.29 ($\pm$0.82) \\
psuedo-label       &  & 80.00 ($\pm$0.50)&79.29 ($\pm$0.96)&79.65 ($\pm$0.97)&79.65 ($\pm$0.96) \\
CGL                &  & 80.04 ($\pm$0.82)&80.19 ($\pm$0.35)&79.75 ($\pm$0.74)&79.75 ($\pm$0.67) \\ \bottomrule
\end{tabular}
\end{table}

\begin{table}[t]
\centering
\caption{\small \textbf{$\Delta_A$ on UTKFace for LBC.}}
\begin{tabular}{cccccc}
\toprule
                  & 100\%             & 80\%              & 50\%              & 25\%              & 10\%              \\ \midrule
group-labeled only & \multirow{4}{*}{18.75 ($\pm$1.04)} & 19.58 ($\pm$2.95)&21.58 ($\pm$1.66)&22.58 ($\pm$1.04)&24.67 ($\pm$2.25) \\
random label       &  & 19.42 ($\pm$0.76)&21.00 ($\pm$0.97)&23.08 ($\pm$0.86)&22.17 ($\pm$1.07) \\
psuedo-label       &  & 19.08 ($\pm$1.16)&19.17 ($\pm$1.17)&19.75 ($\pm$1.93)&19.92 ($\pm$1.99) \\
CGL                &  & 18.00 ($\pm$2.90)&17.92 ($\pm$1.66)&17.83 ($\pm$1.83)&19.25 ($\pm$1.64) \\ \bottomrule
\end{tabular}
\end{table}

\begin{table}[t]
\centering
\caption{\small \textbf{$\Delta_M$ on UTKFace for LBC.}}
\begin{tabular}{cccccc}
\toprule
                  & 100\%            & 80\%              & 50\%              & 25\%              & 10\%               \\ \midrule
group-labeled only & \multirow{4}{*}{33.50 ($\pm$2.69)} & 34.50 ($\pm$3.84)&38.50 ($\pm$1.12)&41.25 ($\pm$3.96)&42.50 ($\pm$7.09) \\
random label       &  & 36.25 ($\pm$1.09)&39.25 ($\pm$2.77)&40.25 ($\pm$1.92)&40.75 ($\pm$2.95) \\
psuedo-label       &  & 33.75 ($\pm$2.17)&33.00 ($\pm$2.00)&35.50 ($\pm$3.35)&36.50 ($\pm$2.87) \\
CGL                &  & 31.50 ($\pm$5.12)&32.25 ($\pm$1.64)&35.00 ($\pm$3.32)&34.00 ($\pm$1.87) \\ \bottomrule
\end{tabular}
\end{table}

\begin{table}[t]
\centering
\caption{\small \textbf{Accuracy on CelebA for MFD.}}
\begin{tabular}{cccccc}
\toprule
                  & 100\%             & 25\%              & 10\%              & 5\%              & 1\%              \\ \midrule
group-labeled only & \multirow{4}{*}{90.14 ($\pm$0.12)} & 89.03 ($\pm$0.28)&88.96 ($\pm$0.49)&87.50 ($\pm$0.42)&82.50 ($\pm$2.08) \\
random label       &  & 88.96 ($\pm$0.07)&87.71 ($\pm$0.21)&87.78 ($\pm$0.69)&86.74 ($\pm$0.07) \\
psuedo-label       &  & 90.49 ($\pm$0.49)&90.69 ($\pm$0.28)&90.62 ($\pm$0.07)&90.62 ($\pm$0.07) \\
CGL                &  & 89.86 ($\pm$0.14)&90.90 ($\pm$0.07)&90.49 ($\pm$0.07)&90.14 ($\pm$0.28) \\ \bottomrule
\end{tabular}
\end{table}

\begin{table}[t]
\centering
\caption{\small \textbf{$\Delta_A$ on CelebA for MFD.}}
\begin{tabular}{cccccc}
\toprule
                  & 100\%             & 25\%              & 10\%              & 5\%              & 1\%              \\ \midrule
group-labeled only & \multirow{4}{*}{5.28 ($\pm$0.69)} & 4.72 ($\pm$0.56)&4.03 ($\pm$0.69)&3.61 ($\pm$0.00)&8.61 ($\pm$0.00) \\
random label       &  & 11.53 ($\pm$0.14)&15.97 ($\pm$0.69)&16.39 ($\pm$0.56)&18.47 ($\pm$0.97) \\
psuedo-label       &  & 5.42 ($\pm$0.69)&5.28 ($\pm$0.56)&5.14 ($\pm$0.42)&6.25 ($\pm$0.14) \\
CGL                &  & 5.28 ($\pm$0.83)&4.03 ($\pm$0.14)&4.58 ($\pm$0.42)&6.39 ($\pm$0.56) \\ \bottomrule
\end{tabular}
\end{table}

\begin{table}[t]
\centering
\caption{\small \textbf{$\Delta_M$ on CelebA for MFD.}}
\begin{tabular}{cccccc}
\toprule
                  & 100\%          & 25\%              & 10\%              & 5\%              & 1\%                 \\ \midrule
group-labeled only & \multirow{4}{*}{8.33 ($\pm$1.04)} & 7.78 ($\pm$1.67)&5.83 ($\pm$0.83)&6.67 ($\pm$0.00)&15.56 ($\pm$0.56) \\
random label       &  & 20.00 ($\pm$0.00)&26.67 ($\pm$2.22)&27.22 ($\pm$1.11)&31.67 ($\pm$1.11) \\
psuedo-label       &  & 9.44 ($\pm$0.00)&10.28 ($\pm$1.39)&9.17 ($\pm$1.39)&11.39 ($\pm$0.28) \\
CGL                &  & 8.06 ($\pm$0.83)&7.22 ($\pm$0.56)&7.78 ($\pm$0.00)&10.83 ($\pm$1.39) \\ \bottomrule
\end{tabular}
\end{table}

\begin{table}[t]
\centering
\caption{\small \textbf{Accuracy on CelebA for FairHSIC.}}
\begin{tabular}{cccccc}
\toprule
                  & 100\%             & 25\%              & 10\%              & 5\%              & 1\%              \\ \midrule
group-labeled only & \multirow{4}{*}{87.22 ($\pm$0.42)} & 83.82 ($\pm$0.07)&81.11 ($\pm$1.39)&80.83 ($\pm$1.94)&74.72 ($\pm$1.25) \\
random label       &  & 84.86 ($\pm$0.14)&85.90 ($\pm$0.21)&84.93 ($\pm$0.35)&85.56 ($\pm$0.14) \\
psuedo-label       &  & 87.99 ($\pm$0.76)&89.31 ($\pm$0.69)&88.19 ($\pm$0.42)&88.82 ($\pm$0.49) \\
CGL                &  & 87.50 ($\pm$1.11)&87.78 ($\pm$1.39)&87.50 ($\pm$1.11)&88.68 ($\pm$0.07) \\ \bottomrule
\end{tabular}
\end{table}

\begin{table}[t]
\centering
\caption{\small \textbf{$\Delta_A$ on CelebA for FairHSIC.}}
\begin{tabular}{cccccc}
\toprule
                  & 100\%             & 25\%              & 10\%              & 5\%              & 1\%              \\ \midrule
group-labeled only & \multirow{4}{*}{12.50 ($\pm$1.11)} & 10.42 ($\pm$3.75)&15.83 ($\pm$2.50)&12.50 ($\pm$3.61)&14.72 ($\pm$2.78) \\
random label       &  & 20.00 ($\pm$1.11)&18.19 ($\pm$0.42)&19.31 ($\pm$0.14)&20.00 ($\pm$0.83) \\
psuedo-label       &  & 10.14 ($\pm$3.19)&9.17 ($\pm$0.28)&12.50 ($\pm$0.28)&7.92 ($\pm$0.97) \\
CGL                &  & 11.67 ($\pm$1.11)&10.28 ($\pm$4.17)&12.22 ($\pm$1.11)&9.31 ($\pm$2.36) \\ \bottomrule
\end{tabular}
\end{table}

\begin{table}[t]
\centering
\caption{\small \textbf{$\Delta_M$ on CelebA for FairHSIC.}}
\begin{tabular}{cccccc}
\toprule
                  & 100\%          & 25\%              & 10\%              & 5\%              & 1\%                 \\ \midrule
group-labeled only & \multirow{4}{*}{20.56 ($\pm$1.67)} & 18.61 ($\pm$6.39)&26.94 ($\pm$3.61)&22.22 ($\pm$6.11)&24.72 ($\pm$1.94) \\
random label       &  & 32.78 ($\pm$2.78)&30.00 ($\pm$1.67)&32.78 ($\pm$0.00)&34.17 ($\pm$1.39) \\
psuedo-label       &  & 17.50 ($\pm$4.72)&13.61 ($\pm$0.83)&20.28 ($\pm$1.39)&13.33 ($\pm$1.67) \\
CGL                &  & 20.28 ($\pm$4.17)&17.22 ($\pm$7.78)&20.00 ($\pm$3.33)&13.61 ($\pm$4.17) \\ \bottomrule
\end{tabular}
\end{table}

\begin{table}[t]
\centering
\caption{\small \textbf{Accuracy on CelebA for LBC.}}
\begin{tabular}{cccccc}
\toprule
                  & 100\%             & 25\%              & 10\%              & 5\%              & 1\%              \\ \midrule
group-labeled only & \multirow{4}{*}{77.57 ($\pm$1.46)} & 73.54 ($\pm$0.63)&74.86 ($\pm$1.25)&76.60 ($\pm$1.18)&72.29 ($\pm$3.12) \\
random label       &  & 78.19 ($\pm$0.14)&78.75 ($\pm$0.28)&79.03 ($\pm$0.56)&78.89 ($\pm$0.14) \\
psuedo-label       &  & 78.06 ($\pm$1.11)&77.57 ($\pm$0.49)&76.39 ($\pm$0.42)&76.25 ($\pm$0.14) \\
CGL                &  & 75.49 ($\pm$0.21)&76.39 ($\pm$0.69)&76.32 ($\pm$0.07)&76.81 ($\pm$1.39) \\ \bottomrule
\end{tabular}
\end{table}

\begin{table}[t]
\centering
\caption{\small \textbf{$\Delta_A$ on CelebA for LBC.}}
\begin{tabular}{cccccc}
\toprule
                  & 100\%             & 25\%              & 10\%              & 5\%              & 1\%              \\ \midrule
group-labeled only & \multirow{4}{*}{12.36 ($\pm$0.14)} & 13.75 ($\pm$2.08)&15.28 ($\pm$3.06)&13.47 ($\pm$0.14)&18.19 ($\pm$2.92) \\
random label       &  & 21.94 ($\pm$0.28)&24.17 ($\pm$1.11)&23.61 ($\pm$1.11)&25.28 ($\pm$0.00) \\
psuedo-label       &  & 12.50 ($\pm$1.39)&13.19 ($\pm$2.92)&11.11 ($\pm$0.28)&10.00 ($\pm$0.56) \\
CGL                &  & 12.08 ($\pm$0.14)&9.17 ($\pm$0.56)&8.47 ($\pm$0.42)&8.33 ($\pm$0.83) \\ \bottomrule
\end{tabular}
\end{table}

\begin{table}[t]
\centering
\caption{\small \textbf{$\Delta_M$ on CelebA for LBC.}}
\begin{tabular}{cccccc}
\toprule
                  & 100\%          & 25\%              & 10\%              & 5\%              & 1\%                 \\ \midrule
group-labeled only & \multirow{4}{*}{23.61 ($\pm$0.83)} & 26.67 ($\pm$3.89)&30.00 ($\pm$5.56)&25.83 ($\pm$0.83)&35.00 ($\pm$5.56) \\
random label       &  & 43.61 ($\pm$0.83)&47.22 ($\pm$2.78)&45.83 ($\pm$3.06)&49.44 ($\pm$0.56) \\
psuedo-label       &  & 24.44 ($\pm$2.78)&26.11 ($\pm$6.11)&21.67 ($\pm$1.11)&19.17 ($\pm$0.83) \\
CGL                &  & 23.89 ($\pm$0.56)&17.50 ($\pm$0.83)&16.39 ($\pm$1.39)&16.39 ($\pm$1.39) \\ \bottomrule
\end{tabular}
\end{table}

\begin{table}[t]
\centering
\caption{\small \textbf{Accuracy on COPMAS for MFD.}}
\begin{tabular}{cccccc}
\toprule
                  & 100\%             & 80\%              & 50\%              & 25\%              & 10\%              \\ \midrule
group-labeled only & \multirow{4}{*}{62.30 ($\pm$0.37)} & 63.61 $(\pm$0.45) & 64.67 $(\pm$0.49) & 62.28 $(\pm$1.33) & 59.95 $(\pm$1.55) \\
random label       &  & 63.15 $(\pm$0.74) & 63.86 $(\pm$0.90) & 64.14 $(\pm$0.70) & 64.87 $(\pm$0.66) \\
psuedo-label       &  & 63.23 $(\pm$0.48) & 64.24 $(\pm$0.78) & 63.61 $(\pm$1.27) & 64.32 $(\pm$0.51) \\
CGL                &  & 63.07 $(\pm$0.68) & 64.08 $(\pm$0.59) & 63.17 $(\pm$0.68) & 63.61 $(\pm$1.22) \\ \bottomrule
\end{tabular}
\end{table}
\begin{table}[t]
\centering
\caption{\small \textbf{$\Delta_A$ on COPMAS for MFD.}}
\begin{tabular}{cccccc}
\toprule
                  & 100\%             & 80\%              & 50\%              & 25\%              & 10\%              \\ \midrule
group-labeled only & \multirow{4}{*}{6.52 ($\pm$0.97)} & 8.57 ($\pm$0.34)&13.59 ($\pm$2.08)&11.72 ($\pm$0.90)&5.13 ($\pm$1.13) \\
random label       &  & 7.57 ($\pm$1.48)&11.72 ($\pm$0.66)&12.84 ($\pm$1.67)&14.15 ($\pm$1.21) \\
psuedo-label       &  & 6.88 ($\pm$0.92)&8.95 ($\pm$1.02)&11.09 ($\pm$1.80)&12.87 ($\pm$1.68) \\
CGL                &  & 6.27 ($\pm$1.08)&7.99 ($\pm$0.65)&10.70 ($\pm$1.90)&10.82 ($\pm$2.18) \\ \bottomrule
\end{tabular}
\end{table}

\begin{table}[t]
\centering
\caption{\small \textbf{$\Delta_M$ on COPMAS for MFD.}}
\begin{tabular}{cccccc}
\toprule
                  & 100\%             & 80\%              & 50\%              & 25\%              & 10\%              \\ \midrule
group-labeled only & \multirow{4}{*}{7.18 ($\pm$0.89)} & 10.24 ($\pm$1.14)&17.13 ($\pm$2.64)&14.96 ($\pm$2.40)&7.15 ($\pm$0.69) \\
random label       &  & 9.67 ($\pm$3.05)&14.86 ($\pm$0.56)&17.13 ($\pm$2.68)&18.39 ($\pm$2.58) \\
psuedo-label       &  & 8.35 ($\pm$1.97)&11.57 ($\pm$0.88)&15.46 ($\pm$2.12)&15.55 ($\pm$2.73) \\
CGL                &  & 7.28 ($\pm$1.66)&10.36 ($\pm$0.54)&14.82 ($\pm$2.60)&13.57 ($\pm$4.15) \\ \bottomrule
\end{tabular}
\end{table}

\begin{table}[t]
\centering
\caption{\small \textbf{Accuracy on COPMAS for FairHSIC.}}
\begin{tabular}{cccccc}
\toprule
                  & 100\%             & 80\%              & 50\%              & 25\%              & 10\%              \\ \midrule
group-labeled only & \multirow{4}{*}{63.94 ($\pm$0.36)} & 64.40 ($\pm$0.70)&64.65 ($\pm$0.31)&62.26 ($\pm$1.12)&58.95 ($\pm$1.46)\\
random label       &  & 64.99 ($\pm$0.24)&64.69 ($\pm$1.18)&64.22 ($\pm$0.66)&63.05 ($\pm$0.94) \\
psuedo-label       &  & 64.83 ($\pm$0.28)&63.17 ($\pm$0.26)&63.53 ($\pm$0.64)&63.82 ($\pm$0.65) \\
CGL                &  & 63.31 ($\pm$0.64)&63.55 ($\pm$0.51)&63.21 ($\pm$0.33)&63.61 ($\pm$0.82) \\ \bottomrule
\end{tabular}
\end{table}

\begin{table}[t]
\centering
\caption{\small \textbf{$\Delta_A$ on COPMAS for FairHSIC.}}
\begin{tabular}{cccccc}
\toprule
                  & 100\%             & 80\%              & 50\%              & 25\%              & 10\%              \\ \midrule
group-labeled only & \multirow{4}{*}{7.63 ($\pm$1.20)} & 9.80 ($\pm$1.21)&11.65 ($\pm$2.14)&11.32 ($\pm$1.16)&6.59 ($\pm$1.90) \\
random label       &  & 11.66 ($\pm$1.25)&11.05 ($\pm$1.88)&11.91 ($\pm$1.90)&11.74 ($\pm$1.49) \\
psuedo-label       &  & 9.92 ($\pm$1.24)&7.76 ($\pm$1.26)&9.91 ($\pm$1.85)&11.57 ($\pm$1.21) \\
CGL                &  & 6.01 ($\pm$1.71)&8.12 ($\pm$1.32)&9.37 ($\pm$2.11)&10.63 ($\pm$1.85) \\ \bottomrule
\end{tabular}
\end{table}

\begin{table}[t]
\centering
\caption{\small \textbf{$\Delta_M$ on COPMAS for FairHSIC.}}
\begin{tabular}{cccccc}
\toprule
                  & 100\%             & 80\%              & 50\%              & 25\%              & 10\%              \\ \midrule
group-labeled only & \multirow{4}{*}{9.66 ($\pm$1.46)} & 11.65 ($\pm$2.01)&14.66 ($\pm$2.37)&14.51 ($\pm$1.73)&9.36 ($\pm$2.67) \\
random label       &  & 14.42 ($\pm$2.78)&14.30 ($\pm$1.39)&16.04 ($\pm$1.78)&15.01 ($\pm$3.32) \\
psuedo-label       &  & 11.91 ($\pm$2.04)&10.56 ($\pm$1.01)&13.07 ($\pm$1.38)&16.51 ($\pm$2.68) \\
CGL                &  & 8.13 ($\pm$3.01)&10.27 ($\pm$1.89)&12.98 ($\pm$2.84)&14.43 ($\pm$3.13) \\ \bottomrule
\end{tabular}
\end{table}

\begin{table}[t]
\centering
\caption{\small \textbf{Accuracy on COPMAS for LBC.}}
\begin{tabular}{cccccc}
\toprule
                  & 100\%             & 80\%              & 50\%              & 25\%              & 10\%              \\ \midrule
group-labeled only & \multirow{4}{*}{61.73 ($\pm$0.12)} & 63.05 ($\pm$0.21)&63.90 ($\pm$0.95)&61.99 ($\pm$1.47)&58.77 ($\pm$1.31) \\
random label       &  & 64.81 ($\pm$0.25)&66.51 ($\pm$0.44)&66.77 ($\pm$0.30)&66.79 ($\pm$0.14) \\
psuedo-label       &  & 63.09 ($\pm$0.90)&65.36 ($\pm$0.27)&66.07 ($\pm$0.39)&66.11 ($\pm$0.93) \\
CGL                &  & 63.01 ($\pm$0.83)&64.20 ($\pm$1.41)&65.70 ($\pm$0.28)&65.80 ($\pm$1.08) \\ \bottomrule
\end{tabular}
\end{table}

\begin{table}[t]
\centering
\caption{\small \textbf{$\Delta_A$ on COPMAS for LBC.}}
\begin{tabular}{cccccc}
\toprule
                  & 100\%             & 80\%              & 50\%              & 25\%              & 10\%              \\ \midrule
group-labeled only & \multirow{4}{*}{4.36 ($\pm$0.69)} & 6.05 ($\pm$1.37)&8.94 ($\pm$1.72)&11.31 ($\pm$0.42)&7.61 ($\pm$0.60) \\
random label       &  & 9.01 ($\pm$0.99)&14.39 ($\pm$1.28)&17.70 ($\pm$0.74)&18.93 ($\pm$0.56) \\
psuedo-label       &  & 5.59 ($\pm$1.28)&11.20 ($\pm$0.91)&14.70 ($\pm$1.53)&16.80 ($\pm$1.04) \\
CGL                &  & 4.99 ($\pm$1.48)&10.32 ($\pm$1.91)&14.24 ($\pm$0.74)&15.56 ($\pm$1.63)\\ \bottomrule
\end{tabular}
\end{table}

\begin{table}[t]
\centering
\caption{\small \textbf{$\Delta_M$ on COPMAS for LBC.}}
\label{tab:deltaM-Co-L}
\begin{tabular}{cccccc}
\toprule
                  & 100\%             & 80\%              & 50\%              & 25\%              & 10\%              \\ \midrule
group-labeled only & \multirow{4}{*}{7.30 ($\pm$1.04)} & 8.18 ($\pm$1.57)&11.63 ($\pm$1.92)&14.33 ($\pm$1.44)&11.02 ($\pm$2.31) \\
random label       &  & 11.94 ($\pm$1.32)&17.99 ($\pm$1.79)&21.71 ($\pm$0.98)&22.91 ($\pm$1.21) \\
psuedo-label       &  & 8.79 ($\pm$1.59)&14.49 ($\pm$1.44)&18.40 ($\pm$1.68)&20.02 ($\pm$2.46) \\
CGL                &  & 7.83 ($\pm$2.35)&13.21 ($\pm$2.91)&18.24 ($\pm$0.42)&18.85 ($\pm$2.50) \\ \bottomrule
\end{tabular}
\end{table}

\clearpage

\end{document}